\documentclass{article}

\PassOptionsToPackage{numbers, compress}{natbib}

    \usepackage[final]{neurips_2019}

\usepackage[utf8]{inputenc}
\usepackage[T1]{fontenc}
\usepackage[colorlinks=true]{hyperref}
\usepackage{booktabs}
\usepackage{amsfonts}
\usepackage{nicefrac}
\usepackage{microtype}
\usepackage{xcolor}
\usepackage{amsmath}
\usepackage{amssymb}
\usepackage{graphicx}
\usepackage{comment}
\usepackage{tikz}
\usepackage{footmisc}
\usetikzlibrary{shapes, arrows, shadows, fit, backgrounds, calc}
\usepackage{bm}
\usepackage{multirow}
\usepackage[titletoc,title]{appendix}
\usepackage{subcaption}
\graphicspath{ {./fig/} }

\newcommand{\method}{BigBiGAN}
\newcommand{\trainval}{train$_{\mathrm{val}}$}
\DeclareMathOperator{\supp}{supp}

\title{Large Scale Adversarial Representation Learning}

\author{%
  Jeff Donahue \\
  DeepMind \\
  \texttt{jeffdonahue@google.com} \\
  \And
  Karen Simonyan \\
  DeepMind \\
  \texttt{simonyan@google.com} \\
}

\begin{document}

\maketitle

\begin{abstract}
Adversarially trained generative models (GANs) have recently achieved compelling image synthesis results.
But despite early successes in using GANs for unsupervised representation learning, they have since been superseded by approaches based on self-supervision.
In this work we show that progress in image generation quality translates to substantially improved representation learning performance.
Our approach,~\method{}, builds upon the state-of-the-art BigGAN model, extending it to representation learning by adding an encoder and modifying the discriminator.
We extensively evaluate the representation learning and generation capabilities of these \method{} models,
demonstrating that these generation-based models achieve the state of the art in unsupervised representation learning on ImageNet, as well as in unconditional image generation.
Pretrained~\method{} models -- including image generators and encoders -- are available on TensorFlow Hub%
\footnote{\label{hubfootnote}
Models available at
\url{https://tfhub.dev/s?publisher=deepmind&q=bigbigan},
with a Colab notebook demo at
\url{https://colab.research.google.com/github/tensorflow/hub/blob/master/examples/colab/bigbigan_with_tf_hub.ipynb}.%
}.

\end{abstract}

\section{Introduction}

In recent years we have seen rapid progress in generative models of visual data.
While these models were previously confined to domains with single or few modes, simple structure, and low resolution,
with advances in both modeling and hardware
they have since gained the ability to convincingly generate
complex, multimodal, high resolution image distributions~\cite{biggan,stylegan,glow}.

Intuitively, the ability to generate data in a particular domain necessitates
a high-level understanding of the semantics of said domain.
This idea has long-standing appeal as raw data is both cheap
-- readily available in virtually infinite supply from sources like the Internet --
and rich, with images comprising far more information than the class labels
that typical discriminative machine learning models are trained to predict from them.
Yet, while the progress in generative models has been undeniable,
nagging questions persist:
what semantics have these models learned,
and how can they be leveraged for representation learning?

The dream of generation as a means of true understanding from raw data alone has hardly been realized.
Instead, the most successful approaches for unsupervised learning leverage techniques adopted from the field of supervised learning,
a class of methods known as \textit{self-supervised} learning~\cite{carl,splitbrain,cpc,rotation}.
These approaches typically involve changing or holding back certain aspects of the data in some way,
and training a model to predict or generate aspects of the missing information.
For example,~\cite{colorful,splitbrain} proposed colorization as a means of unsupervised learning,
where a model is given a subset of the color channels in an input image, and trained to predict the missing channels.

Generative models as a means of unsupervised learning offer
an appealing alternative to self-supervised tasks
in that they are trained to model the full data distribution
without requiring any modification of the original data.
One class of generative models that has been applied to representation learning is generative adversarial networks (GANs)~\cite{gan}.
The generator in the GAN framework is a feed-forward mapping from randomly sampled latent variables (also called ``noise'') to generated data,
with learning signal provided by a \textit{discriminator} trained to distinguish between real and generated data samples,
guiding the generator's outputs to follow the data distribution.
The \textit{adversarially learned inference} (ALI)~\cite{ali} or \textit{bidirectional GAN} (BiGAN)~\cite{bigan} approaches
were proposed as extensions to the GAN framework that augment the standard GAN with an \textit{encoder} module mapping real data to latents,
the inverse of the mapping learned by the generator.

In the limit of an optimal discriminator,~\cite{bigan} showed that a deterministic BiGAN behaves like an autoencoder minimizing $\ell_0$ reconstruction costs;
however, the shape of the reconstruction error surface 
is dictated by a parametric discriminator,
as opposed to simple pixel-level measures like the $\ell_2$ error.
Since the discriminator is usually a powerful neural network, the hope is that it will induce
an error surface which emphasizes ``semantic'' errors in reconstructions, rather than low-level details.

In~\cite{bigan} it was demonstrated that the encoder learned via the BiGAN or ALI framework is an effective
means of visual representation learning on ImageNet for downstream tasks.
However, it used a DCGAN~\cite{dcgan} style generator, incapable of producing high-quality images on this dataset,
so the semantics the encoder could model were in turn quite limited.
In this work we revisit this approach using BigGAN~\cite{biggan} as the generator,
a modern model that appears capable of capturing many of the modes
and much of the structure present in ImageNet images.
Our contributions are as follows:
\begin{itemize}
\item We show that BigBiGAN (BiGAN with BigGAN generator) matches the state of the art in unsupervised representation learning on ImageNet.
\item We propose a more stable version of the joint discriminator for \method.
\item We perform a thorough empirical analysis and ablation study of model design choices.
\item We show that the representation learning objective also improves unconditional image generation, and demonstrate state-of-the-art results in unconditional ImageNet generation.
\item We open source pretrained \method{} models on TensorFlow Hub\footnote{See footnote \footref{hubfootnote}.}.
\end{itemize}

\section{\method{}}
\label{sec:method}

\newcommand{\loss}{\mathcal{L}}
\newcommand{\disc}{\mathcal{D}}
\newcommand{\gen}{\mathcal{G}}
\newcommand{\enc}{\mathcal{E}}
\newcommand{\x}{\mathbf{x}}
\newcommand{\z}{\mathbf{z}}
\newcommand{\score}{s}

\newcommand{\statcirc}[2]{%
    \fill[#2] (0,0) circle (0.5cm);

    \fill[#1] (0,0) -- (45:0.5cm) arc (45:225:0.5cm) -- cycle;
}

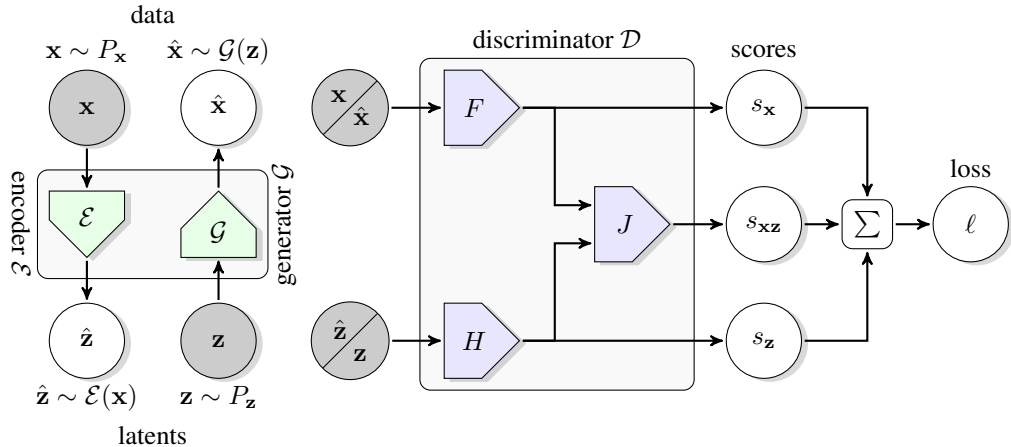
\begin{figure}[t]

\tikzstyle{shadow} = [copy shadow={opacity=0.15, shadow xshift=0.5ex,shadow yshift=-0.25ex,fill=black}]
\tikzstyle{all_nodes} = [minimum width=1cm, minimum height=0.5cm,text centered, draw=black,fill=white,shadow]
\tikzstyle{observed_var} = [circle, all_nodes, fill=black!20]
\tikzstyle{shadow2} = [copy shadow={opacity=0.15, shadow xshift=0.5ex,shadow yshift=-0.25ex,fill=black}]
\tikzstyle{observed_var_split} = [forbidden sign, fill=black!20, minimum width=1cm, minimum height=1cm,text centered, draw=black, opacity=1.]
\tikzstyle{latent_var} = [circle, all_nodes]
\tikzstyle{group_var} = [ellipse, all_nodes, inner sep=0.1cm]
\tikzstyle{point} = [circle]
\tikzstyle{func} = [signal, signal to=east, all_nodes,minimum height=1cm,fill=red!10]
\tikzstyle{discfunc} = [signal, signal to=east, all_nodes,minimum height=1cm,fill=blue!10]
\tikzstyle{minigenfunc} = [signal, signal to=north, all_nodes, minimum height=0.3cm, minimum width=0.5cm, fill=green!10]
\tikzstyle{miniencfunc} = [signal, signal to=south, all_nodes, minimum height=0.3cm, minimum width=0.5cm, fill=green!10]
\tikzstyle{encfunc} = [signal, signal to=south, all_nodes, minimum height=0.5cm, minimum width=1cm, fill=green!10]
\tikzstyle{genfunc} = [signal, signal to=north, all_nodes, minimum height=0.5cm, minimum width=1cm, fill=green!10]
\tikzstyle{arrow}= [->,thick,shorten <=1pt,shorten >=1pt, >=stealth']
\tikzstyle{box}= [rectangle, rounded corners, draw=black, inner sep=0.15cm]
\tikzstyle{widebox}= [rectangle, rounded corners, draw=black, inner sep=0.15cm]
\tikzset{
  -|-/.style={
    to path={
      (\tikztostart) -| ($(\tikztostart)!#1!(\tikztotarget)$) |- (\tikztotarget)
      \tikztonodes
    }
  },
  -|-/.default=0.7,
}

\newif\ifShowEGInD
\newif\ifShowFHOuts
\newif\ifShowEGLeft

\ShowEGLefttrue

\newcommand{\lsp}{1.5}

\ifShowFHOuts
  \newcommand{\fhd}{1.75}
\else
  \newcommand{\fhd}{0.}
\fi
\ifShowEGLeft

  \newcommand{\egld}{0.3}
\else
  \newcommand{\egld}{0.}
\fi

\centering
\scalebox{1.0}{
 \begin{tikzpicture}[node distance=2cm]
  \ifShowEGLeft

   \node (x1) [observed_var] at (-5.25,1.25+\egld) {$\x$};
   \node (Eleft) [encfunc] at (-5.25,0.25) {};
   \node (Elefttext) at (-5.25,0.1) {$\enc$};
   \draw[arrow] (x1) -> (Eleft);
   \node (z1) [latent_var] at (-5.25,-1.25-\egld) {};
   \node (z1text) at (-5.25,-1.2-\egld) {$\hat{\z}$};
   \draw[arrow] (Eleft) -> (z1);
   \node[rotate=270] (Elefttext) at (-6.125,0) {encoder $\enc$};

   \node (z2) [observed_var] at (-3.5,-1.25-\egld) {$\z$};
   \node (Gleft) [genfunc] at (-3.5,-0.25) {};
   \node (Glefttext) at (-3.5,-0.1) {$\gen$};
   \draw[arrow] (z2) -> (Gleft);
   \node (x2) [latent_var] at (-3.5,1.25+\egld) {};
   \node (x2text) at (-3.5,1.3+\egld) {$\hat{\x}$};
   \draw[arrow] (Gleft) -> (x2);
   \node[rotate=90] (Glefttext) at (-2.625,0) {generator $\gen$};

   \node (extlefttop) [point] at (-4.125,0.4) {};
   \node (extleftbottom) [point] at (-4.125,-0.4) {};

   \begin{scope}[on background layer]
    \node[widebox,fill=gray!5,fit=(Eleft) (Gleft) (extlefttop) (extleftbottom)] (container) {};
   \end{scope}
  \fi

  \node (x) [observed_var_split] at (-1.75,1.25+\egld) {\raisebox{0.15cm}{$\x$} \raisebox{-0.15cm}{$\hat{\x}$}};
  \node (z) [observed_var_split] at (-1.75,-1.25-\egld) {\raisebox{0.15cm}{$\hat{\z}$} \raisebox{-0.15cm}{$\z$}};
  
  \ifShowEGLeft

   \node (xtext1) at (-5.25,2+\egld) {$\x \sim P_\x$};
   \node (xtext2) at (-3.5,2+\egld) {$\hat{\x} \sim \gen(\z)$};
   \node (xtext3) at (-4.375,2.5+\egld) {data};
   \node (ztext1) at (-3.5,-2-\egld) {$\z \sim P_\z$};
   \node (ztext2) at (-5.25,-2-\egld) {$\hat{\z} \sim \enc(\x)$};
   \node (ztext3) at (-4.375,-2.5-\egld) {latents};
  \else
   \node (xtext) at (-1.75,2+\egld) {data $\x \sim P_\x$};
   \node (ztext) at (-1.75,-2-\egld) {latents $\z \sim P_\z$};
  \fi

  \ifShowEGInD
    \node (G) [minigenfunc] at (-1.375,-0.125) {$\gen$};
    \draw[arrow] (z.north east) -> (G);
    \draw[arrow] (G) -> (x.south east);

    \node (E) [miniencfunc] at (-2.125,0.125) {$\enc$};
    \draw[arrow] (x.south west) -> (E);
    \draw[arrow] (E) -> (z.north west);
  \fi

  \node (extleft) [point] at (-0.5,0) {};

  \node (F) [discfunc] at (-0.25,1.25+\egld) {};
  \node (Ftext) at (-0.1,1.25+\egld) {$F$};
  \draw[arrow] (x) -> (F);
  \node (H) [discfunc] at (-0.25,-1.25-\egld) {};
  \node (Htext) at (-0.1,-1.25-\egld) {$H$};
  \draw[arrow] (z) -> (H);

  \ifShowFHOuts
   \node (Fx) [latent_var] at (1.75,1.25+\egld) {$F(\x)$};
   \draw[arrow] (F) -> (Fx);
   \node (Hz) [latent_var] at (1.75,-1.25-\egld) {$H(\z)$};
   \draw[arrow] (H) -> (Hz);
  \fi

  \node (J) [discfunc] at (1.75+\fhd,0) {};
  \node (Jtext) at (1.9+\fhd,0) {$J$};
  \ifShowFHOuts
   \draw[arrow] (Hz) to[-|-] ($(J.south west)!0.5!(J.west)$);
   \draw[arrow] (Fx) to[-|-] ($(J.north west)!0.5!(J.west)$);
  \else
   \draw[arrow] (F) to[-|-] ($(J.north west)!0.5!(J.west)$);
   \draw[arrow] (H) to[-|-] ($(J.south west)!0.5!(J.west)$);
  \fi

  \node (extright) [point] at (2.5+\fhd,0) {};

  \begin{scope}[on background layer]
    \node[widebox,fill=gray!5,fit=(extleft) (F) (H) (J) (extright), label={above:discriminator $\disc$}] (container) {};
  \end{scope}

  \node (scorestext) at (3.75+\fhd,2.0+\egld) {scores};
  \node (sx) [latent_var] at (3.75+\fhd,1.25+\egld) {$\score_\x$};
  \node (sz) [latent_var] at (3.75+\fhd,-1.25-\egld) {$\score_\z$};
  \node (sxz) [latent_var] at (3.75+\fhd,0) {$\score_{\x\z}$};
  \draw[arrow] (J) -> (sxz);
  \ifShowFHOuts
   \draw[arrow] (Fx) -> (sx);
   \draw[arrow] (Hz) -> (sz);
  \else
   \draw[arrow] (F) -> (sx);
   \draw[arrow] (H) -> (sz);
  \fi

  \node (scoresum) [box] at (5.125+\fhd,0) {$\sum$};
  \draw[arrow] (sx) -| (scoresum);
  \draw[arrow] (sxz) -> (scoresum);
  \draw[arrow] (sz) -| (scoresum);

  \node (losstext) at (6.5+\fhd,0.75) {loss};
  \node (L) [latent_var] at (6.5+\fhd,0) {$\ell$};
  \draw[arrow] (scoresum) -> (L);

 \end{tikzpicture}
}

\caption{
The structure of the \method{} framework.
The joint discriminator $\disc$ is used to compute the loss $\ell$.
Its inputs are data-latent pairs, either $(\x \sim P_\x, \hat{\z} \sim \enc(\x))$, sampled from the data distribution $P_\x$ and encoder $\enc$ outputs, or $(\hat{\x} \sim \gen(\z), \z \sim P_\z)$, sampled from the generator $\gen$ outputs and the latent distribution $P_\z$.
The loss $\ell$ includes the unary data term $\score_\x$ and the unary latent term $\score_\z$, as well as the joint term $\score_{\x\z}$ which ties the data and latent distributions.
}
\label{fig:joint_discrim}
\end{figure}

The BiGAN~\cite{bigan} or ALI~\cite{ali} approaches were proposed as extensions of the GAN~\cite{gan} framework which enable the learning of an encoder that can be employed as an inference model~\cite{ali} or feature representation~\cite{bigan}.
Given a distribution $P_\x$ of data $\x$ (e.g., images),
and a distribution $P_\z$ of latents $\z$ (usually a simple continuous distribution like an isotropic Gaussian $\mathcal{N}(0, I)$),
the generator $\gen$ models a conditional distribution $P(\x|\z)$ of data $\x$ given latent inputs $\z$ sampled from the latent prior $P_\z$,
as in the standard GAN generator~\cite{gan}.
The encoder $\enc$ models the inverse conditional distribution $P(\z|\x)$, predicting latents $\z$ given data $\x$ sampled from the data distribution $P_\x$.

Besides the addition of $\enc$, the other modification to the GAN in the BiGAN framework is a joint discriminator $\disc$,
which takes as input data-latent pairs $(\x, \z)$ (rather than just data $\x$ as in a standard GAN),
and learns to discriminate between pairs from the data distribution and encoder, versus the generator and latent distribution.
Concretely, its inputs are pairs
$(\x \sim P_\x, \hat{\z} \sim \enc(\x))$
and
$(\hat{\x} \sim \gen(\z), \z \sim P_\z)$,
and the goal of the $\gen$ and $\enc$ is to ``fool'' the discriminator by making the two
joint distributions $P_{\x\enc}$ and $P_{\gen\z}$ from which these pairs are sampled indistinguishable.
The adversarial minimax objective in~\cite{bigan,ali}, analogous to that of the GAN framework~\cite{gan}, was defined as follows:
\begin{align*}
    \min_{\gen\enc} \max_{\disc} \left\{
        \mathbb{E}_{\x \sim P_\x, \z \sim \enc_\Phi(\x)} \left[ \log(\sigma(\disc(\x, \z))) \right] +
        \mathbb{E}_{\z \sim P_\z, \x \sim \gen_\Phi(\z)} \left[ \log(1-\sigma(\disc(\x, \z))) \right]
    \right\}
\end{align*}
Under this objective,~\cite{bigan,ali} showed that with an optimal $\disc$, $\gen$ and $\enc$ minimize the Jensen-Shannon divergence between the joint distributions $P_{\x\enc}$ and $P_{\gen\z}$, and therefore at the global optimum, the two joint distributions $P_{\x\enc} = P_{\gen\z}$ match, analogous to the results from standard GANs~\cite{gan}.
Furthermore,~\cite{bigan} showed that in the case where $\enc$ and $\gen$ are deterministic functions
(i.e., the learned conditional distributions $P_\gen(\x|\z)$ and $P_\enc(\z|\x)$ are Dirac $\delta$ functions),
these two functions are inverses at the global optimum: e.g.,
$\forall_{\x \in \supp(P_\x)} \, \x = \gen(\enc(\x))$,
with the optimal joint discriminator effectively imposing $\ell_0$ reconstruction costs on $\x$ and $\z$.

While the crux of our approach,~\method{}, remains the same as that of BiGAN~\cite{bigan,ali},
we have adopted the generator and discriminator architectures from the state-of-the-art BigGAN~\cite{biggan} generative image model.
Beyond that, we have found that an improved discriminator structure leads to better representation learning results without compromising generation
(Figure~\ref{fig:joint_discrim}).
Namely, in addition to the joint discriminator loss proposed in~\cite{bigan,ali} which ties the data and latent distributions together,
we propose additional unary terms in the learning objective, which are functions only of either the data $\x$ or the latents $\z$.
Although~\cite{bigan,ali} prove that the original BiGAN objective already enforces that the learnt joint distributions match at the global optimum,
implying that the marginal distributions of $\x$ and $\z$ match as well,
these unary terms intuitively guide optimization in the ``right direction''
by explicitly enforcing this property.
For example, in the context of image generation,
the unary loss term on $\x$ 
matches the original GAN objective and provides a learning signal
which steers only the generator to match the image distribution independently of its latent inputs.
(In our evaluation we will demonstrate empirically that the addition of these terms results in both improved generation and representation learning.)

Concretely, the discriminator loss $\loss_{\disc}$ and the encoder-generator loss $\loss_{\enc\gen}$ are defined as follows, based on scalar discriminator ``score'' functions $s_*$ and the corresponding per-sample losses $\ell_*$:
\begin{align*}
    \score_\x(\x) &=
        \theta_\x^{\intercal} F_\Theta(\x) \\
    \score_\z(\z) &=
        \theta_\z^{\intercal} H_\Theta(\z) \\
    \score_{\x\z}(\x, \z) &=
        \theta_{\x\z}^{\intercal} J_\Theta(F_\Theta(\x), H_\Theta(\z)) \\
    \ell_{\enc\gen}(\x, \z, y) &=
         y \left(\score_\x(\x) + \score_\z(\z) + \score_{\x\z}(\x, \z) \right) & y \in \{-1, +1\} \\
    \loss_{\enc\gen}(P_\x, P_\z) &=
        \mathbb{E}_{\x \sim P_\x, \hat{\z} \sim \enc_\Phi(\x)} \left[ \ell_{\enc\gen}(\x, \hat{\z}, +1) \right] +
        \mathbb{E}_{\z \sim P_\z, \hat{\x} \sim \gen_\Phi(\z)} \left[ \ell_{\enc\gen}(\hat{\x}, \z, -1) \right] \\
    \ell_{\disc}(\x, \z, y) &=
         h(y \score_\x(\x)) + h(y \score_\z(\z)) + h(y \score_{\x\z}(\x, \z)) & y \in \{-1, +1\}  \\
    \loss_{\disc}(P_\x, P_\z) &=
        \mathbb{E}_{\x \sim P_\x, \hat{\z} \sim \enc_\Phi(\x)} \left[ \ell_{\disc}(\x, \hat{\z}, +1) \right] +
        \mathbb{E}_{\z \sim P_\z, \hat{\x} \sim \gen_\Phi(\z)} \left[ \ell_{\disc}(\hat{\x}, \z, -1) \right]
\end{align*} %
where $h(t) = \max(0, 1-t)$ is a ``hinge'' used to regularize the discriminator~\cite{geometricgan,tran}
\footnote{
  We also considered an alternative discriminator loss $\ell'_{\disc}$ which invokes the ``hinge'' $h$ just once on the sum of the three loss terms -- $
    \ell'_{\disc}(\x, \z, y) =
         h(y \left( \score_\x(\x) + \score_\z(\z) + \score_{\x\z}(\x, \z) \right))
  $ -- but found that this performed significantly worse than $\ell_{\disc}$ above which clamps each of the three loss terms separately.
},
also used in BigGAN~\cite{biggan}.
The discriminator $\disc$ includes three submodules: $F$, $H$, and $J$.
$F$ takes only $\x$ as input and $H$ takes only $\z$, and learned projections of their outputs with parameters $\theta_\x$ and $\theta_\z$ respectively give the scalar unary scores $\score_\x$ and $\score_\z$.
In our experiments, the data $\x$ are images and latents $\z$ are unstructured flat vectors; accordingly, $F$ is a ConvNet and $H$ is an MLP.
The joint score $\score_{\x\z}$ tying $\x$ and $\z$ is given by the remaining $\disc$ submodule, $J$, a function of the outputs of $F$ and $H$.

The $\enc$ and $\gen$ parameters $\Phi$ are optimized to minimize the loss $\loss_{\enc\gen}$,
and the $\disc$ parameters $\Theta$ are optimized to minimize loss $\loss_{\disc}$.
As usual, the expectations $\mathbb{E}$ are estimated by Monte Carlo samples taken over minibatches.

Like in BiGAN~\cite{bigan} and ALI~\cite{ali},
the discriminator loss $\loss_{\disc}$ intuitively trains the discriminator to distinguish between the two joint data-latent distributions from the encoder and the generator,
pushing it to predict positive values for encoder input pairs $(\x, \enc(\x))$ and negative values for generator input pairs $(\gen(\z), \z)$.
The generator and encoder loss $\loss_{\enc\gen}$ trains these two modules to fool the discriminator into incorrectly predicting the opposite,
in effect pushing them to create matching joint data-latent distributions.
(In the case of deterministic $\enc$ and $\gen$, this requires the two modules to invert one another~\cite{bigan}.)

\section{Evaluation}
\label{sec:eval}

Most of our experiments follow the standard protocol used to evaluate unsupervised learning techniques,
first proposed in~\cite{colorful}.
We train a \method{} on unlabeled ImageNet,
freeze its learned representation,
and then train a linear classifier on its outputs,
fully supervised using all of the training set labels.
We also measure image generation performance,
reporting Inception Score~\cite{improvedgan} (IS) and Fr\'echet Inception Distance~\cite{frechet} (FID)
as the standard metrics there.

\subsection{Ablation}
We begin with an extensive ablation study in which we directly evaluate a number of modeling choices,
with results presented in Table~\ref{ablations}.
Where possible we performed three runs of each variant with different seeds and report the mean and standard deviation for each metric.

We start with a relatively fully-fledged version of the model at $128\times128$ resolution (row \textit{Base}),
with the $\gen$ architecture and the $F$ component of $\disc$ taken from the corresponding $128 \times 128$ architectures in BigGAN, including the skip connections and shared noise embedding proposed in~\cite{biggan}.
$\z$ is 120 dimensions, split into six groups of 20 dimensions fed into each of the six layers of $\gen$ as in~\cite{biggan}.
The remaining components of $\disc$ -- $H$ and $J$ -- are 8-layer MLPs with ResNet-style skip connections (four residual blocks with two layers each) and size 2048 hidden layers.
The $\enc$ architecture is the ResNet-v2-50 ConvNet originally proposed for image classification in~\cite{resnetv2}, followed by a 4-layer MLP (size 4096) with skip connections (two residual blocks) after ResNet's globally average pooled output.
The unconditional BigGAN training setup corresponds to the ``Single Label'' setup proposed in~\cite{zurichfewer}, where a single ``dummy'' label is used for all images (theoretically equivalent to learning a bias in place of the class-conditional batch norm inputs).
We then ablate several aspects of the model, with results detailed in the following paragraphs.
Additional architectural and optimization details are provided in Appendix~\ref{appendix_model_details}.
Full learning curves for many results are included in Appendix~\ref{appendix_curves}.

\paragraph{Latent distribution $P_\z$ and stochastic $\enc$.}
As in ALI~\cite{ali}, the encoder $\enc$ of our \textit{Base} model is non-deterministic,
parametrizing a distribution $\mathcal{N(\mu, \sigma)}$.
$\mu$ and $\hat{\sigma}$ are given by a linear layer at the output of the model,
and the final standard deviation $\sigma$ is computed from $\hat{\sigma}$
using a non-negative ``softplus'' non-linearity $\sigma = \log(1 + \exp(\hat{\sigma}))$~\cite{softplus}.
The final $\z$ uses the reparametrized sampling from~\cite{kingmavae}, with $\z = \mu + \epsilon \sigma$, where $\epsilon \sim \mathcal{N}(0, I)$.
Compared to a deterministic encoder (row \textit{Deterministic $\enc$}) which predicts $\z$ directly without sampling
(effectively modeling $P(\z|\x)$ as a Dirac $\delta$ distribution), the non-deterministic \textit{Base} model achieves significantly better classification performance
(at no cost to generation).
We also compared to using a uniform $P_\z = \mathcal{U}(-1, 1)$ (row \textit{Uniform $P_\z$})
with $\enc$ deterministically predicting $\z = \tanh({\hat{\z}})$ given a linear output $\hat{\z}$, as done in BiGAN~\cite{bigan}.
This also achieves worse classification results than the non-deterministic \textit{Base} model.

\paragraph{Unary loss terms.}
We evaluate the effect of removing one or both unary terms of the loss function proposed in Section~\ref{sec:method}, $s_{\x}$ and $s_{\z}$.
Removing both unary terms (row \textit{No Unaries}) corresponds to the original objective proposed in~\cite{bigan,ali}.
It is clear that the $\x$ unary term has a large positive effect on generation performance,
with the \textit{Base} and \textit{$\x$ Unary Only} rows having significantly better IS and FID
than the \textit{$\z$ Unary Only} and \textit{No Unaries} rows.
This result makes intuitive sense as it matches the standard generator loss.
It also marginally improves classification performance.
The $\z$ unary term makes a more marginal difference,
likely due to the relative ease of modeling relatively simple distributions like isotropic Gaussians,
though also does result in slightly improved classification and generation in terms of FID --
especially without the $\x$ term (\textit{$\z$ Unary Only} vs. \textit{No Unaries}).
On the other hand, IS is worse with the $\z$ term.
This may be due to IS roughly measuring the generator's coverage of the major modes of the distribution (the classes)
rather than the distribution in its entirety,
the latter of which may be better captured by FID and more likely to be promoted by a good encoder $\enc$.
The requirement of invertibility in a (Big)BiGAN could be encouraging the generator to produce distinguishable outputs across the entire latent space, rather than ``collapsing'' large volumes of latent space to a single mode of the data distribution.

\paragraph{$\gen$ capacity.}
To address the question of the importance of the generator $\gen$ in representation learning,
we vary the capacity of $\gen$ (with $\enc$ and $\disc$ fixed) in the \textit{Small $\gen$} rows.
With a third of the capacity of the \textit{Base} $\gen$ model (\textit{Small $\gen$ (32)}),
the overall model is quite unstable and achieves significantly worse classification results than the higher capacity base model%
\footnote{Though the generation performance by IS and FID in row \textit{Small $\gen$ (32)} is very poor at the point we measured -- when its best validation classification performance (43.59\%) is achieved --
this model was performing more reasonably for generation earlier in training, reaching IS 14.69 and FID 60.67.}
With two-thirds capacity (\textit{Small $\gen$ (64)}), generation performance is substantially worse (matching the results in~\cite{biggan})
and classification performance is modestly worse.
These results confirm that a powerful image generator is indeed important for learning good representations via the encoder.
Assuming this relationship holds in the future, we expect that better generative models are likely to lead to further improvements in representation learning.

\paragraph{Standard GAN.}
We also compare \method{}'s image generation performance against a standard unconditional BigGAN with no encoder $\enc$ and only the standard $F$ ConvNet in the discriminator,
with only the $s_\x$ term in the loss
(row \textit{No $\enc$ (GAN)}).
While the standard GAN achieves a marginally better IS, the \method{} FID is about the same,
indicating that the addition of the \method{} $\enc$ and joint $\disc$ does not compromise generation with the newly proposed unary loss terms described in Section~\ref{sec:method}.
(In comparison, the versions of the model without unary loss term on $\x$ -- rows \textit{$\z$ Unary Only} and \textit{No Unaries} --
have substantially worse generation performance in terms of FID than the standard GAN.)
We conjecture that the IS is worse for similar reasons that the $s_\z$ unary loss term leads to worse IS.
Next we will show that with an enhanced $\enc$ taking higher input resolutions, generation with \method{} in terms of FID is substantially improved over the standard GAN.

\paragraph{High resolution $\enc$ with varying resolution $\gen$.}
BiGAN~\cite{bigan} proposed an asymmetric setup in which $\enc$ takes higher resolution images than $\gen$ outputs and $\disc$ takes as input, showing that an $\enc$ taking $128\times128$ inputs with a $64\times64$ $\gen$ outperforms a $64\times64$ $\enc$ for downstream tasks.
We experiment with this setup in \method{}, raising the $\enc$ input resolution to $256 \times 256$
-- matching the resolution used in typical supervised ImageNet classification setups --
and varying the $\gen$ output and $\disc$ input resolution in $\{64, 128, 256\}$.
Our results in Table~\ref{ablations} (rows \textit{High Res $\enc$ (256)} and \textit{Low/High Res $\gen$ (*)}) show that \method{} achieves better representation learning results as the $\gen$ resolution increases, up to the full $\enc$ resolution of $256\times256$.
However, because the overall model is much slower to train with $\gen$ at $256\times256$ resolution, the remainder of our results use the $128\times128$ resolution for $\gen$.

Interestingly, with the higher resolution $\enc$, generation improves significantly (especially by FID), despite $\gen$ operating at the same resolution (row \textit{High Res $\enc$ (256)} vs. \textit{Base}).
This is an encouraging result for the potential of \method{} as a means of improving adversarial image synthesis itself, besides its use in representation learning and inference.

\paragraph{$\enc$ architecture.}
Keeping the $\enc$ input resolution fixed at 256, we experiment with varied and often larger $\enc$ architectures, including several of the ResNet-50 variants explored in~\cite{revisiting}. In particular, we expand the capacity of the hidden layers by a factor of $2$ or $4$, as well as swap the residual block structure to a reversible variant called \textit{RevNet}~\cite{revnet} with the same number of layers and capacity as the corresponding ResNets.
(We use the version of RevNet described in~\cite{revisiting}.)
We find that the base ResNet-50 model (row \textit{High Res $\enc$ (256)}) outperforms RevNet-50 (row \textit{RevNet}),
but as the network widths are expanded, we begin to see improvements from RevNet-50, with double-width RevNet outperforming a ResNet of the same capacity (rows \textit{RevNet $\times 2$} and \textit{ResNet $\times 2$}).
We see further gains with an even larger quadruple-width RevNet model (row \textit{RevNet $\times 4$}), which we use for our final results in Section~\ref{sec:priorcomp}.

\paragraph{Decoupled $\enc$/$\gen$ optimization.}
As a final improvement,
we decoupled the $\enc$ optimizer from that of $\gen$,
and found that simply using a $10 \times$ higher learning rate for $\enc$ dramatically accelerates training
and improves final representation learning results.
For ResNet-50 this improves linear classifier accuracy by nearly 3\% (\textit{ResNet ($\uparrow \enc$ LR)} vs. \textit{High Res $\enc$ (256)}).
We also applied this to our largest $\enc$ architecture, RevNet-50 $\times 4$,
and saw similar gains (\textit{RevNet $\times 4$ ($\uparrow \enc$ LR)} vs. \textit{RevNet $\times 4$}).

\newcommand{\yes}{\checkmark}
\newcommand{\no}{-}
\newcommand{\resnet}{S}
\newcommand{\revnet}{V}
\newcommand{\phz}{\hphantom{0}}

\newcommand{\ch}[1]{{\color{blue} (#1)}}
\newcommand{\chb}[1]{\ch{#1}}

\setlength{\tabcolsep}{2pt}
\begin{table*}
\centering
\footnotesize
\scalebox{0.85}{
\begin{tabular}{l|cccccc|cc|ccc|c|ccc}
\toprule
  & \multicolumn{6}{|c|}{Encoder ($\enc$)} & \multicolumn{2}{|c|}{Gen. ($\gen$)} & \multicolumn{3}{|c|}{Loss $\loss_*$} & & \multicolumn{3}{|c}{Results}\\
  & A. & D. & C. & R. & Var. & $\eta$ & C. & R. & $\score_{\x\z}$ & $\score_\x$ & $\score_\z$ & $P_\z$ & IS ($\uparrow$) & FID ($\downarrow$) & Cls. ($\uparrow$) \\
\midrule
   Base                      & \resnet & 50 & 1 & 128 & \yes & 1 & 96 & 128 & \yes & \yes & \yes & $\mathcal{N}$ & 22.66 $\pm$ 0.18   & 31.19 $\pm$ 0.37   & 48.10 $\pm$ 0.13 \\
 \hline
   Deterministic $\enc$      & \resnet & 50 & 1 & 128 & \ch{\no}  & 1 & 96 & 128 & \yes & \yes & \yes & $\mathcal{N}$ & 22.79 $\pm$ 0.27   & 31.31 $\pm$ 0.30   & 46.97 $\pm$ 0.35 \\
   Uniform $P_\z$            & \resnet & 50 & 1 & 128 & \ch{\no}  & 1 & 96 & 128 & \yes & \yes & \yes & \ch{$\mathcal{U}$} & 22.83 $\pm$ 0.24   & 31.52 $\pm$ 0.28   & 45.11 $\pm$ 0.93 \\
 \hline
   $\x$ Unary Only           & \resnet & 50 & 1 & 128 & \yes & 1 & 96 & 128 & \yes & \yes & \ch{\no}  & $\mathcal{N}$ & 23.19 $\pm$ 0.28   & 31.99 $\pm$ 0.30   & 47.74 $\pm$ 0.20 \\
   $\z$ Unary Only           & \resnet & 50 & 1 & 128 & \yes & 1 & 96 & 128 & \yes & \ch{\no}  & \yes & $\mathcal{N}$ & 19.52 $\pm$ 0.39   & 39.48 $\pm$ 1.00   & 47.78 $\pm$ 0.28 \\
   No Unaries (BiGAN)        & \resnet & 50 & 1 & 128 & \yes & 1 & 96 & 128 & \yes & \ch{\no}  & \ch{\no}  & $\mathcal{N}$ & 19.70 $\pm$ 0.30   & 42.92 $\pm$ 0.92   & 46.71 $\pm$ 0.88 \\
 \hline
   Small $\gen$ (32)         & \resnet & 50 & 1 & 128 & \yes & 1 & \ch{32} & 128 & \yes & \yes & \yes & $\mathcal{N}$ &  3.28 $\pm$ 0.18   & 247.30 $\pm$ 10.31 & 43.59 $\pm$ 0.34 \\
   Small $\gen$ (64)         & \resnet & 50 & 1 & 128 & \yes & 1 & \ch{64} & 128 & \yes & \yes & \yes & $\mathcal{N}$ & 19.96 $\pm$ 0.15   & 38.93 $\pm$ 0.39   & 47.54 $\pm$ 0.33 \\
 \hline

   No $\enc$ (GAN) *         & \multicolumn{6}{|c|}{\ch{\no}}    & 96 & 128 & \ch{\no}  & \yes & \ch{\no} & $\mathcal{N}$ & 23.56 $\pm$ 0.37   & 30.91 $\pm$ 0.23 & - \\
\hline
\hline
  High Res $\enc$ (256) & \resnet & 50 & 1 & \ch{256} & \yes & 1 & 96 & 128 & \yes & \yes & \yes & $\mathcal{N}$ & 23.45 $\pm$ 0.14   & 27.86 $\pm$ 0.13 & 50.80 $\pm$ 0.30 \\
\hline
  Low Res $\gen$ (64) & \resnet & 50 & 1 & \chb{256} & \yes & 1 & 96 &  \ch{64} & \yes & \yes & \yes & $\mathcal{N}$ & 19.40 $\pm$ 0.19   & 15.82 $\pm$ 0.06 & 47.51 $\pm$ 0.09 \\

  High Res $\gen$ (256) & \resnet & 50 & 1 & \chb{256} & \yes & 1 & 96 & \ch{256} & \yes & \yes & \yes & $\mathcal{N}$ & 24.70              & 38.58            & 51.49            \\
\hline
  ResNet-101                 & \resnet & \ch{101} & 1 & \chb{256} & \yes & 1 & 96 & 128 & \yes & \yes & \yes & $\mathcal{N}$ & 23.29 & 28.01 & 51.21 \\
  ResNet $\times 2$          & \resnet & 50 & \ch{2} & \chb{256} & \yes & 1 & 96 & 128 & \yes & \yes & \yes & $\mathcal{N}$ & 23.68              & 27.81            & 52.66 \\
  RevNet                     & \ch{\revnet} & 50 & 1 & \chb{256} & \yes & 1 & 96 & 128 & \yes & \yes & \yes & $\mathcal{N}$ & 23.33 $\pm$ 0.09   & 27.78 $\pm$ 0.06 & 49.42 $\pm$ 0.18 \\
  RevNet $\times 2$          & \ch{\revnet} & 50 & \ch{2} & \chb{256} & \yes & 1 & 96 & 128 & \yes & \yes & \yes & $\mathcal{N}$ & 23.21              & 27.96            & 54.40 \\
  RevNet $\times 4$          & \ch{\revnet} & 50 & \ch{4} & \chb{256} & \yes & 1 & 96 & 128 & \yes & \yes & \yes & $\mathcal{N}$ & 23.23              & 28.15            & 57.15 \\
\hline
  ResNet ($\uparrow \enc$ LR)& \resnet & 50 & 1 & \ch{256} & \yes & \ch{10} & 96 & 128 & \yes & \yes & \yes & $\mathcal{N}$ & 23.27 $\pm$ 0.22   & 28.51 $\pm$ 0.44 & 53.70 $\pm$ 0.15 \\
  RevNet $\times 4$ ($\uparrow \enc$ LR)         & \ch{\revnet} & 50 & \ch{4} & \chb{256} & \yes & \ch{10} & 96 & 128 & \yes & \yes & \yes & $\mathcal{N}$ & 23.08 & 28.54 & 60.15 \\
\bottomrule
\end{tabular}
}
 \caption{
  Results for variants of \method{}, given in Inception Score~\cite{improvedgan} (IS) and Fr\'echet Inception Distance~\cite{frechet} (FID) of the generated images,
  and ImageNet top-1 classification accuracy percentage (Cls.) of a supervised logistic regression classifier trained on the encoder features~\cite{colorful}, computed on a split of 10K images randomly sampled from the training set, which we refer to as the ``\trainval{}'' split.
  The \textit{Encoder ($\enc$)} columns specify the $\enc$ architecture (A.) as ResNet (\resnet{}) or RevNet (\revnet{}),
  the depth (D., e.g. 50 for ResNet-50),
  the channel width multiplier (C.), with 1 denoting the original widths from~\cite{resnetv2},
  the input image resolution (R.),
  whether the variance is predicted and a $\z$ vector is sampled from the resulting distribution (Var.),
  and the learning rate multiplier $\eta$ relative to the $\gen$ learning rate.
  The \textit{Generator ($\gen$)} columns specify the BigGAN $\gen$ channel multiplier (C.),
  with 96 corresponding to the original width from~\cite{biggan},
  and output image resolution (R.).
  The \textit{Loss} columns specify which terms of the \method{} loss are present in the objective.
  The $P_z$ column specifies the input distribution as a standard normal $\mathcal{N}(0, 1)$ or continuous uniform $\mathcal{U}(-1, 1)$.
  Changes from the \textit{Base} setup in each row are {\color{blue} highlighted in blue}.
  Results with margins of error (written as ``$\mu \pm \sigma$'') are the means and standard deviations over three runs with different random seeds.
  (Experiments requiring more computation were run only once.)
  (* Result for vanilla GAN (\textit{No $\enc$ (GAN)}) selected with early stopping based on best FID; other results selected with early stopping based on validation classification accuracy (Cls.).)
 }
 \label{ablations}
\end{table*}
\setlength{\tabcolsep}{6pt}

\subsection{Comparison with prior methods}
\label{sec:priorcomp}

\paragraph{Representation learning.}
\begin{table*}
\centering
\scalebox{1.0}{
\begin{tabular}{lll|cc}
\toprule
 Method & Architecture & Feature & Top-1 & Top-5 \\
\midrule
 BiGAN~\cite{bigan,splitbrain}                  & AlexNet      & Conv3    & 31.0 & -    \\
 SS-GAN~\cite{ssgan}                            & ResNet-19    & Block6   & 38.3 & -    \\
 Motion Segmentation (MS)~\cite{motionseg,carl} & ResNet-101   & AvePool  & 27.6 & 48.3 \\
 Exemplar (Ex)~\cite{exemplar,carl}             & ResNet-101   & AvePool  & 31.5 & 53.1 \\
 Relative Position (RP)~\cite{carlorig,carl}    & ResNet-101   & AvePool  & 36.2 & 59.2 \\
 Colorization (Col)~\cite{colorful,carl}        & ResNet-101   & AvePool  & 39.6 & 62.5 \\
 Combination of MS+Ex+RP+Col~\cite{carl}        & ResNet-101   & AvePool  & -    & 69.3 \\
 CPC~\cite{cpc}                                 & ResNet-101   & AvePool  & 48.7 & 73.6 \\
 Rotation~\cite{rotation,revisiting}            & RevNet-50 $\times 4$ & AvePool  & 55.4 & -    \\
 Efficient CPC~\cite{cpcplusplus}               & ResNet-170   & AvePool  & 61.0 & 83.0 \\
 \hline

 \multirow{4}{*}{\method{} (ours)}

                                                & ResNet-50    & AvePool  & 55.4 & 77.4 \\
                                                & ResNet-50    & BN+CReLU & 56.6 & 78.6 \\
                                                & RevNet-50 $\times 4$  & AvePool  & 60.8 & 81.4 \\
                                                & RevNet-50 $\times 4$  & BN+CReLU & 61.3 & 81.9 \\
\bottomrule
\end{tabular}
}
 \caption{
  Comparison of \method{} models on the official ImageNet validation set
  against recent competing approaches with a supervised logistic regression classifier.
  \method{} results are selected with early stopping based on highest accuracy on our \trainval{} subset of 10K training set images.
  \textit{ResNet-50} results correspond to row \textit{ResNet ($\uparrow \enc$ LR)} in Table~\ref{ablations},
  and
  \textit{RevNet-50 $\times 4$} corresponds to \textit{RevNet $\times 4$ ($\uparrow \enc$ LR)}.
 }
 \label{final_results}
\end{table*}

We now take our best model by \trainval{} classification accuracy
from the above ablations and present results on the official ImageNet validation set,
comparing against the state of the art in recent unsupervised learning literature.
For comparison, we also present classification results for our best performing variant with the smaller ResNet-50-based $\enc$.
These models correspond to the last two rows of Table~\ref{ablations},
\textit{ResNet ($\uparrow \enc$ LR)} and
\textit{RevNet $\times 4$ ($\uparrow \enc$ LR)}.

Results are presented in Table~\ref{final_results}.
(For reference, the fully supervised accuracy of these architectures is given in
Appendix~\ref{appendix_model_details}, Table~\ref{sup_model_perf}.)
Compared with a number of modern self-supervised approaches~\cite{motionseg,carlorig,colorful,cpc,rotation,cpcplusplus}
and combinations thereof~\cite{carl}, our \method{} approach based purely on generative models performs well for representation learning,
state-of-the-art among recent unsupervised learning results,
improving upon a recently published result from~\cite{revisiting} of 55.4\% to 60.8\% top-1 accuracy using rotation prediction pre-training with the same representation learning architecture
\footnote{Our RevNet $\times 4$ architecture matches the widest architectures used in~\cite{revisiting}, labeled as $\times 16$ there.}
and feature, labeled as \textit{AvePool} in Table~\ref{final_results},
and matches the results of the concurrent work in~\cite{cpcplusplus} based on contrastic predictive coding (CPC).

We also experiment with learning linear classifiers on a different rendering of the \textit{AvePool} feature, labeled \textit{BN+CReLU},
which boosts our best results with RevNet $\times 4$ to 61.3\% top-1 accuracy.
Given the global average pooling output $a$, we first compute $h = \mathrm{BatchNorm}(a)$,
and the final feature is computed by concatenating $[\mathrm{ReLU}(h), \mathrm{ReLU}(-h)]$,
sometimes called a ``CReLU'' (concatened ReLU) non-linearity~\cite{crelu}.
$\mathrm{BatchNorm}$ denotes parameter-free Batch Normalization~\cite{batchnorm},
where the scale ($\gamma$) and offset ($\beta$) parameters are not learned,
so training a linear classifier on this feature does not involve any additional learning.
The CReLU non-linearity retains all the information in its inputs and doubles the feature dimension,
each of which likely contributes to the improved results.

Finally, in Appendix~\ref{appendix_neighbors} we consider evaluating representations by zero-shot $k$ nearest neighbors classification,
achieving 43.3\% top-1 accuracy in this setting.
Qualitative examples of nearest neighbors are presented in Figure~\ref{fig:neighbors1}.

\paragraph{Unsupervised image generation.}
\begin{table*}
\centering
\scalebox{0.82}{
 \begin{tabular}{lc|ccc}
 \toprule
  Method & Steps & IS ($\uparrow$) & FID vs. Train ($\downarrow$) & FID vs. Val. ($\downarrow$) \\
 \midrule
  BigGAN + SL~\cite{zurichfewer}  & 500K & 20.4 (15.4 $\pm$ 7.57)    & - & 25.3 (71.7 $\pm$ 66.32) \\
  BigGAN + Clustering~\cite{zurichfewer}    & 500K &22.7 (22.8 $\pm$ 0.42)    & - & 23.2 (22.7 $\pm$ 0.80) \\
  \hline

  \method{} + SL (ours)  & 500K & 25.38 (25.33 $\pm$ 0.17) & 22.78 (22.63 $\pm$ 0.23) & 23.60 (23.56 $\pm$ 0.12) \\
  \method{} High Res $\enc$ + SL (ours)  & 500K & 25.43 (25.45 $\pm$ 0.04) & 22.34 (22.36 $\pm$ 0.04)  & 22.94 (23.00 $\pm$ 0.15) \\
  \hline
  \method{} High Res $\enc$ + SL (ours)  & 1M & 27.94 (27.80 $\pm$ 0.21) & 20.32 (20.27 $\pm$ 0.09) & 21.61 (21.62 $\pm$ 0.09) \\
 \bottomrule
 \end{tabular}
}
 \caption{
  Comparison of our \method{} for unsupervised (unconditional) generation vs. previously reported results for unsupervised BigGAN from~\cite{zurichfewer}.
  We specify the ``pseudo-labeling'' method as \textit{SL} (Single Label) or \textit{Clustering}.
  For comparison we train~\method{} for the same number of steps (500K) as the BigGAN-based approaches from~\cite{zurichfewer},
  but also present results from additional training to 1M steps in the last row and observe further improvements.
  All results above include the median $m$ as well as the mean $\mu$ and standard deviation $\sigma$ across three runs, written as ``$m$ ($\mu$ $\pm$ $\sigma$)''.
  The \method{} result is selected with early stopping based on best FID vs. Train.
 }
 \label{uncond_gen_results}
\end{table*}

In Table~\ref{uncond_gen_results} we show results for unsupervised generation with \method{}, comparing to the BigGAN-based~\cite{biggan} unsupervised generation results from~\cite{zurichfewer}.
Note that these results differ from those in Table~\ref{ablations} due to the use of the data augmentation method of~\cite{zurichfewer}\footnote{See the ``distorted'' preprocessing method from the Compare GAN framework: \url{https://github.com/google/compare_gan/blob/master/compare_gan/datasets.py}.}
(rather than ResNet-style preprocessing used for all results in our Table~\ref{ablations} ablation study).
The lighter augmentation from~\cite{zurichfewer} results in better image generation performance under the IS and FID metrics.
The improvements are likely due in part to the fact that this augmentation, on average, crops larger portions of the image,
thus yielding generators that typically produce images encompassing most or all of a given object,
which tends to result in more representative samples of any given class (giving better IS)
and more closely matching the statistics of full center crops (as used in the real data statistics to compute FID).
Besides this preprocessing difference,
the approaches in Table~\ref{uncond_gen_results} have the same configurations as used in the \textit{Base} or \textit{High Res $\enc$ (256)} row of Table~\ref{ablations}.

These results show that \method{} significantly improves both IS and FID over the baseline unconditional BigGAN generation results with the same (unsupervised) ``labels'' (a single fixed label in the \textit{SL} (Single Label) approach -- row \textit{BigBiGAN + SL} vs. \textit{BigGAN + SL}).
We see further improvements using a high resolution $\enc$ (row \textit{BigBiGAN High Res $\enc$ + SL}), surpassing the previous unsupervised state of the art (row \textit{BigGAN + Clustering}) under both IS and FID.
(Note that the image generation results remain comparable: the generated image resolution is still $128 \times 128$ here, despite the higher resolution $\enc$ input.)
The alternative ``pseudo-labeling'' approach from~\cite{zurichfewer}, \textit{Clustering},
which uses labels derived from unsupervised clustering,
is complementary to \method{} and combining both could yield further improvements.
Finally, observing that results continue to improve significantly with training beyond 500K steps,
we also report results at 1M steps in the final row of Table~\ref{uncond_gen_results}.

\subsection{Reconstruction}
\label{sec:recons}

As shown in~\cite{bigan,ali}, the (Big)BiGAN $\enc$ and $\gen$ can reconstruct data instances $\x$ by computing the encoder's predicted latent representation $\enc(\x)$ and then passing this predicted latent back through the generator to obtain the reconstruction $\gen(\enc(\x))$.
We present \method{} reconstructions in Figure~\ref{fig:recons}.
These reconstructions are far from pixel-perfect, likely due in part to the fact that no reconstruction cost is explicitly enforced by the objective -- reconstructions are not even computed at training time.
However, they may provide some intuition for what features the encoder $\enc$ learns to model.
For example, when the input image contains a dog, person, or a food item, the reconstruction is often a different instance of the same ``category'' with similar pose, position, and texture -- for example, a similar species of dog facing the same direction.
The extent to which these reconstructions tend to retain the high-level semantics of the inputs rather than the low-level details suggests that \method{} training encourages the encoder to model the former more so than the latter.
Additional reconstructions are presented in Appendix~\ref{appendix_samples}.

\newcommand{\reconwidth}{0.13\textwidth}
\begin{figure}
\centering
 \includegraphics[width=\reconwidth]{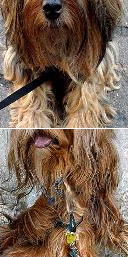}
 \includegraphics[width=\reconwidth]{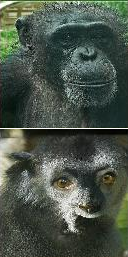}
 \includegraphics[width=\reconwidth]{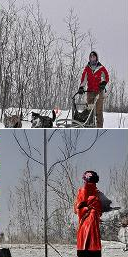}
 \includegraphics[width=\reconwidth]{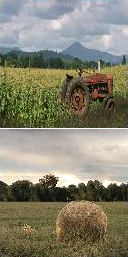}
 \includegraphics[width=\reconwidth]{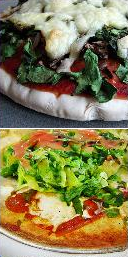}
 \includegraphics[width=\reconwidth]{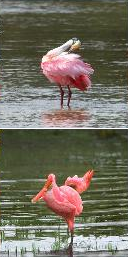}
 \includegraphics[width=\reconwidth]{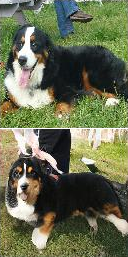}
 \caption{
  Selected reconstructions from an unsupervised \method{} model (Section~\ref{sec:recons}).
  Top row images are real data $\x \sim P_\x$; bottom row images are generated reconstructions of the above image $\x$ computed by $\gen(\enc(\x))$.
  Unlike most explicit reconstruction costs (e.g., pixel-wise), the reconstruction cost implicitly minimized by a (Big)BiGAN~\cite{bigan,ali} tends to emphasize more semantic, high-level details.
  Additional reconstructions are presented in Appendix~\ref{appendix_samples}.
}
 \label{fig:recons}
\end{figure}

\section{Related work}
A number of approaches to unsupervised representation learning from images based on self-supervision have proven very successful.
Self-supervision generally involves learning from tasks designed to resemble supervised learning in some way, but in which the ``labels'' can be created automatically from the data itself with no manual effort.
An early example is relative location prediction~\citep{carlorig}, where a model is trained on input pairs of image patches and predicts their relative locations.
Contrastive predictive coding (CPC)~\citep{cpc,cpcplusplus} is a recent related approach where, given an image patch, a model predicts
which patches occur in other image locations.
Other approaches include colorization~\cite{colorful,splitbrain},
motion segmentation~\cite{motionseg},
rotation prediction~\cite{rotation,ssgan},
GAN-based discrimination~\cite{dcgan,ssgan},
and exemplar matching~\cite{exemplar}.
Rigorous empirical comparisons of many of these approaches have also been conducted~\cite{carl,revisiting}.
A key advantage offered by \method{} and other approaches based on generative models, relative to most self-supervised approaches,
is that their input may be the full-resolution image or other signal, with no cropping or modification of the data needed
(though such modifications may be beneficial as data augmentation).
This means the resulting representation can typically be applied directly to full data in the downstream task with no domain shift.

A number of relevant autoencoder and GAN variants have also been proposed.
Associative compression networks (ACNs)~\cite{acn} learn to compress at the dataset level by conditioning data on other previously transmitted data which are similar in code space, resulting in models that can ``daydream'' semantically similar samples, similar to \method{} reconstructions.
VQ-VAEs~\cite{vqvae} pair a discrete (vector quantized) encoder with an autoregressive decoder to produce faithful reconstructions with a high compression factor and demonstrate representation learning results in reinforcement learning settings.
In the adversarial space, adversarial autoencoders~\cite{advae} proposed an autoencoder-style encoder-decoder pair trained with pixel-level reconstruction cost, replacing the KL-divergence regularization of the prior used in VAEs~\cite{kingmavae} with a discriminator.
In another proposed VAE-GAN hybrid~\cite{learnedsimilarity} the pixel-space reconstruction error used in most VAEs is replaced with feature space distance from an intermediate layer of a GAN discriminator.
Other hybrid approaches like AGE~\cite{age} and $\alpha$-GAN~\cite{alphagan} add an encoder to stabilize GAN training.
An interesting difference between many of these approaches and the BiGAN~\cite{ali,bigan} framework is that BiGAN does not train the encoder or generator with an explicit reconstruction cost.
Though it can be shown that (Big)BiGAN implicitly minimizes a reconstruction cost,
qualitative reconstruction results (Section~\ref{sec:recons}) suggest that this reconstruction cost is of a different flavor, emphasizing high-level semantics over pixel-level details.

\section{Discussion}
We have shown that \method{}, an unsupervised learning approach based purely on generative models, achieves state-of-the-art results in image representation learning on ImageNet.
Our ablation study lends further credence to the hope that powerful generative models can be beneficial for representation learning, and in turn that learning an inference model can improve large-scale generative models.
In the future we hope that representation learning can continue to benefit from further advances in generative models and inference models alike, as well as scaling to larger image databases.

\subsubsection*{Acknowledgments}
The authors would like to thank
    Aidan Clark,
    Olivier H\'{e}naff,
    A\"aron van den Oord,
    Sander Dieleman,
and many other colleagues at DeepMind for
useful discussions and feedback on this work.

\small

\bibliography{references}
\bibliographystyle{plain}

\normalsize

\begin{appendices}

\newpage
\section{Model and optimization details}
\label{appendix_model_details}
Our optimizer matches that of BigGAN~\cite{biggan} -- we use Adam~\cite{adam} with batch size 2048 and the same learning rates and other hyperparameters,
using the $\gen$ optimizer to update $\enc$ simultaneously,
with the same alternating optimization: two $\disc$ updates followed by a single joint update of $\gen$ and $\enc$.
(We do not use orthogonal regularization used in~\cite{biggan}, finding it gave worse results in the unconditional setting, matching the findings of~\cite{zurichfewer}.)
Spectral normalization~\cite{sngan} is used in $\gen$ and $\disc$, but not in $\enc$.
Full cross-replica batch normalization is used in both $\gen$ and $\enc$
(including for the linear classifier training on $\enc$ features used for evaluations).
We also apply exponential moving averaging (EMA) with a decay of 0.9999 to the $\gen$ and $\enc$ weights in all evaluations.
(We find this results in only a small improvement for $\enc$ evaluations, but a substantial one for $\gen$ evaluations.)

At \method{} training time, as well as linear classification evaluation training time, we preprocess inputs with ResNet~\cite{resnetv2}-style data augmentation, though with crops of size 128 or 256 rather than 224\footnote{Preprocessing code from the TensorFlow ResNet TPU model: \url{https://github.com/tensorflow/tpu/tree/master/models/official/resnet}.}.

For linear classification evaluations in the ablations reported in Table~\ref{ablations}, we hold out 10K randomly selected images from the official ImageNet~\cite{imagenet} training set as a validation set and report accuracy on that validation set, which we call \trainval{}.
All results in Table~\ref{ablations} are run for 500K steps, with early stopping based on linear classifier accuracy on our \trainval{} split.
In all of these models the linear classifier is initialized to 0 and trained for 5K Adam steps with a (high) learning rate of 0.01 and EMA smoothing with decay 0.9999.
We have found it helpful to monitor representation learning progress during \method{} training by periodically
rerunning this linear classification evaluation from scratch given the current $\enc$ weights,
resetting the classifier weights to 0 before each evaluation.

In Table~\ref{final_results} we extend the \method{} training time to 1M steps,
and report results on the official validation set of 50K images for comparison with prior work.
The classifier in these results is trained for 100K Adam steps, sweeping over learning rates $\{10^{-4}, 3 \cdot 10^{-4}, 10^{-3}, 3 \cdot 10^{-3}, 10^{-2}\}$, again applying EMA with decay 0.9999 to the classifier weights.
Hyperparameter selection and early stopping is again based on classification accuracy on \trainval{}.
As in~\cite{biggan}, FID is reported against statistics over the full ImageNet training set, preprocessed by resizing the minor axis to the $\gen$ output resolution and taking the center crop along the major axis, except as noted in Table~\ref{uncond_gen_results}, where we also report FID against the validation set for comparison with~\cite{zurichfewer}.

All models were trained
via TensorFlow~\cite{tensorflow} and Sonnet~\cite{sonnet}
with data parallelism on TPU pod slices~\cite{tpu} using 32 to 512 cores,
coordinated by TF-Replicator~\cite{replicator}.

\paragraph{Supervised model performance.}
In Table~\ref{sup_model_perf} we present the results of fully supervised training with the model architectures used in our experiments in Section~\ref{sec:eval} for comparison purposes.

\begin{table*}[h]
\centering
\scalebox{1}{
\begin{tabular}{l|cc}
\toprule
 Architecture & Top-1 & Top-5 \\
\midrule
 ResNet-50            & 76.3 & 93.1 \\
 ResNet-101           & 77.8 & 93.8 \\
 RevNet-50            & 71.8 & 90.5 \\
 RevNet-50 $\times 2$ & 74.9 & 92.2 \\
 RevNet-50 $\times 4$ & 76.6 & 93.1 \\
\bottomrule
\end{tabular}
}
 \caption{
  ImageNet validation set accuracy for fully supervised end-to-end training of the model architectures used in our representation learning experiments.
 }
 \label{sup_model_perf}
\end{table*}

\begin{figure}
 \centering
 \begin{subfigure}[t]{0.45\textwidth}
  \centering
   \includegraphics[width=1.\linewidth]{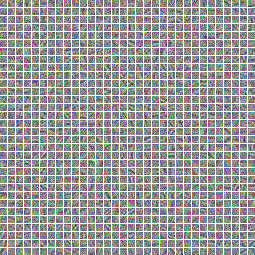}
  \caption{RevNet $\times 4$}
 \end{subfigure}
 \hspace{1cm}
 \begin{subfigure}[t]{0.45\textwidth}
  \centering
   \includegraphics[width=1.\linewidth]{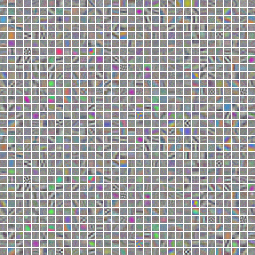}
  \caption{+ $\uparrow \enc$ LR}
 \end{subfigure}
 \caption{Visualization of first layer convolutional filters for our unsupervised \method{} models with the RevNet $\times 4$ $\enc$ architecture, which includes 1024 filters. (Best viewed with zoom.)}
 \label{fig:filters}
\end{figure}

\paragraph{First layer convolutional filters.}
In Figure~\ref{fig:filters} we visualize the learned convolutional filters for the first convolutional layer of our \method{} encoders $\enc$
using the largest RevNet $\times 4$ $\enc$ architecture.
Note the difference between the filters in (a) and (b) 
(corresponding to rows
\textit{RevNet $\times 4$}
and
\textit{RevNet $\times 4$ ($\uparrow \enc$ LR)}
in Table~\ref{ablations}).
In (b) we use the higher $\enc$ learning rate and see a corresponding qualitative improvement in the appearance of the learned filters,
with less noise and more Gabor-like and color filters, as observed in BiGAN~\cite{bigan}.
This suggests that examining the convolutional filters of the input layer can serve as a diagnostic for undertrained models.

\newpage
\section{Samples and reconstructions}
\label{appendix_samples}
\begin{table*}
\centering
\scalebox{1}{
\begin{tabular}{l|ccc|cc}
\toprule
  & \multicolumn{3}{|c|}{Samples} & \multicolumn{2}{|c}{Reconstructions} \\
 Model & Image & IS ($\uparrow$) & FID ($\downarrow$) & Image & Rel. $\ell_1$ Error \% ($\downarrow$) \\
\midrule
  Base
      & Figure~\ref{fig:samples_128} & 24.10 & 30.14
      & Figure~\ref{fig:more_recons_128} & 70.54
  \\
  Light Augmentation
      & Figure~\ref{fig:samples_i128lightaug} & 27.09 & 20.96
      & Figure~\ref{fig:recons_i128lightaug} & 72.53
  \\
  High Res $\enc$ (256) 
      & Figure~\ref{fig:samples_highrese} & 24.91 & 26.56
      & Figure~\ref{fig:recons_highrese} & 70.60
  \\
  High Res $\gen$ (256)
      & Figure~\ref{fig:samples_256} & 25.73 & 37.21
      & Figure~\ref{fig:recons_256} & 77.70
  \\
\bottomrule
\end{tabular}
}
 \caption{
  Links to \method{} samples and reconstructions with associated metrics.
 }
 \label{sample_info}
\end{table*}

In this Appendix we present \method{} samples and reconstructions from several variants of the method.
Table~\ref{sample_info} includes pointers to samples and reconstruction images, as well as relevant metrics.
The samples were selected by best FID vs. training set statistics,
and we show the IS and FID along with sample images at that point.
The reconstructions were selected by best (lowest) relative pixel-wise $\ell_1$ error, the error metric presented in Table~\ref{sample_info}, computed as:
\begin{align*}
 E_{\mathrm{Rel} \ell_1} &=
  \frac{
    \mathbb{E}_{\x \sim P_\x}
    || \x - \gen(\enc(\x)) ||_1
  }{
    \mathbb{E}_{\x, \x' \sim P_\x}
    || \x' - \gen(\enc(\x)) ||_1
  },
\end{align*}
where $\x$ and $\x'$ are independent data samples, and $
    || \x' - \gen(\enc(\x)) ||_1
$ serves as a ``baseline'' reconstruction error relative to a ``random'' input.
For example, with a random initialization of $\gen$ and $\enc$, we have $
E_{\mathrm{Rel} \ell_1} \approx 1
$.
This relative metric penalizes degenerate reconstructions, such as the mean image, which would sometimes achieve low absolute reconstruction error despite having no perceptual similarity to the inputs.
despite that the resulting images having no perceptual similarity to the inputs.
In practice, given $N$ data samples $\x_0, \x_1, \ldots, \x_{N-1}$ (we use $N =$ 50K),
we estimate the denominator by comparing each sample $\x_i$ with a single neighbor $\x_{(i+1) \,\mathrm{mod}\, N}$, computing:
\begin{align*}
 E_{\mathrm{Rel} \ell_1} &\approx
  \frac{
    \sum_{i=0}^{N-1}
    || \x_i - \gen(\enc(\x_i)) ||_1
  }{
    \sum_{i=0}^{N-1}
    || \x_{(i+1) \,\mathrm{mod}\, N} - \gen(\enc(\x_i)) ||_1
  }
\end{align*}

\begin{figure}
\centering
\begin{tabular}{cccccccc}
 \includegraphics[trim=0 2560 0    0,clip,width=0.14\textwidth]{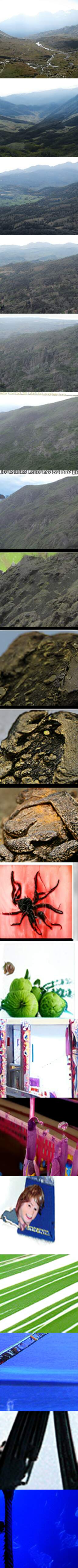} &
 \includegraphics[trim=0    0 0 2560,clip,width=0.14\textwidth]{resnet_ims90pctres_2_qual65.jpg} &
 &
 \includegraphics[trim=0 2560 0    0,clip,width=0.14\textwidth]{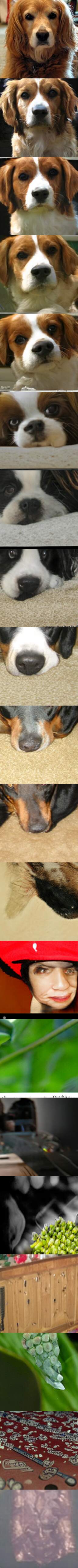} &
 \includegraphics[trim=0    0 0 2560,clip,width=0.14\textwidth]{resnet_ims90pctres_3_qual65.jpg} &
 &
 \includegraphics[trim=0 2560 0    0,clip,width=0.14\textwidth]{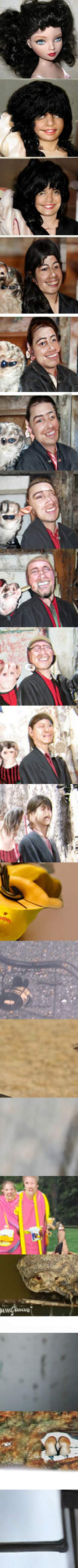} &
 \includegraphics[trim=0    0 0 2560,clip,width=0.14\textwidth]{resnet_ims90pctres_11_qual65.jpg} \\
 $R_0 : R_9$ & $R_{10} : R_{500}$ & &
 $R_0 : R_9$ & $R_{10} : R_{500}$ & &
 $R_0 : R_9$ & $R_{10} : R_{500}$ \\
 \multicolumn{2}{c}{Image 1} & &
 \multicolumn{2}{c}{Image 2} & &
 \multicolumn{2}{c}{Image 3} \\
\end{tabular}
 \caption{
  Iterated reconstructions from an unsupervised \method{} model, trained using the \textit{ResNet ($\uparrow \enc$ LR)} method from Table~\ref{ablations},
  computed by recursively running the reconstruction operation $\gen(\enc(\cdot))$ on its own output as described in Appendix~\ref{appendix_samples}.
  In each pair of columns, the left column shows a real input image $R_0$ at the top, and $R_1$ through $R_9$ in the remaining rows,
  the results of iterating reconstruction one to nine times,
  The right column shows the result of up to 500 iterations sampled at longer intervals, displaying
  $R_{10}$,
  $R_{20}$,
  $R_{30}$,
  $R_{40}$,
  $R_{50}$,
  $R_{100}$,
  $R_{200}$,
  $R_{300}$,
  $R_{400}$, and
  $R_{500}$.
}
 \label{fig:iterrecons}
\end{figure}

\paragraph{Iterated reconstruction}
To further explore the behavior of a BigBiGAN (or any other model capable of approximately reconstructing its input),
we can ``iterate'' the reconstruction operation.
In particular, let $R_i(\x)$ be defined for non-negative integers $i$ and input images $\x$ as:
\begin{align*}
R_0(\x) &= \x \\
R_{i+1}(\x) &= \gen(\enc(R_i(\x)))
\end{align*}
In Figure~\ref{fig:iterrecons} we show the results of up to 500 steps of this process for a few sample images.
Qualitatively, the first several steps of this process often appear to retain some semantics of the input image $\x$.
After dozens or hundreds of iterations, however, little content from the original input apparently remains intact.

\begin{figure}
\centering
 \includegraphics[width=1\textwidth]{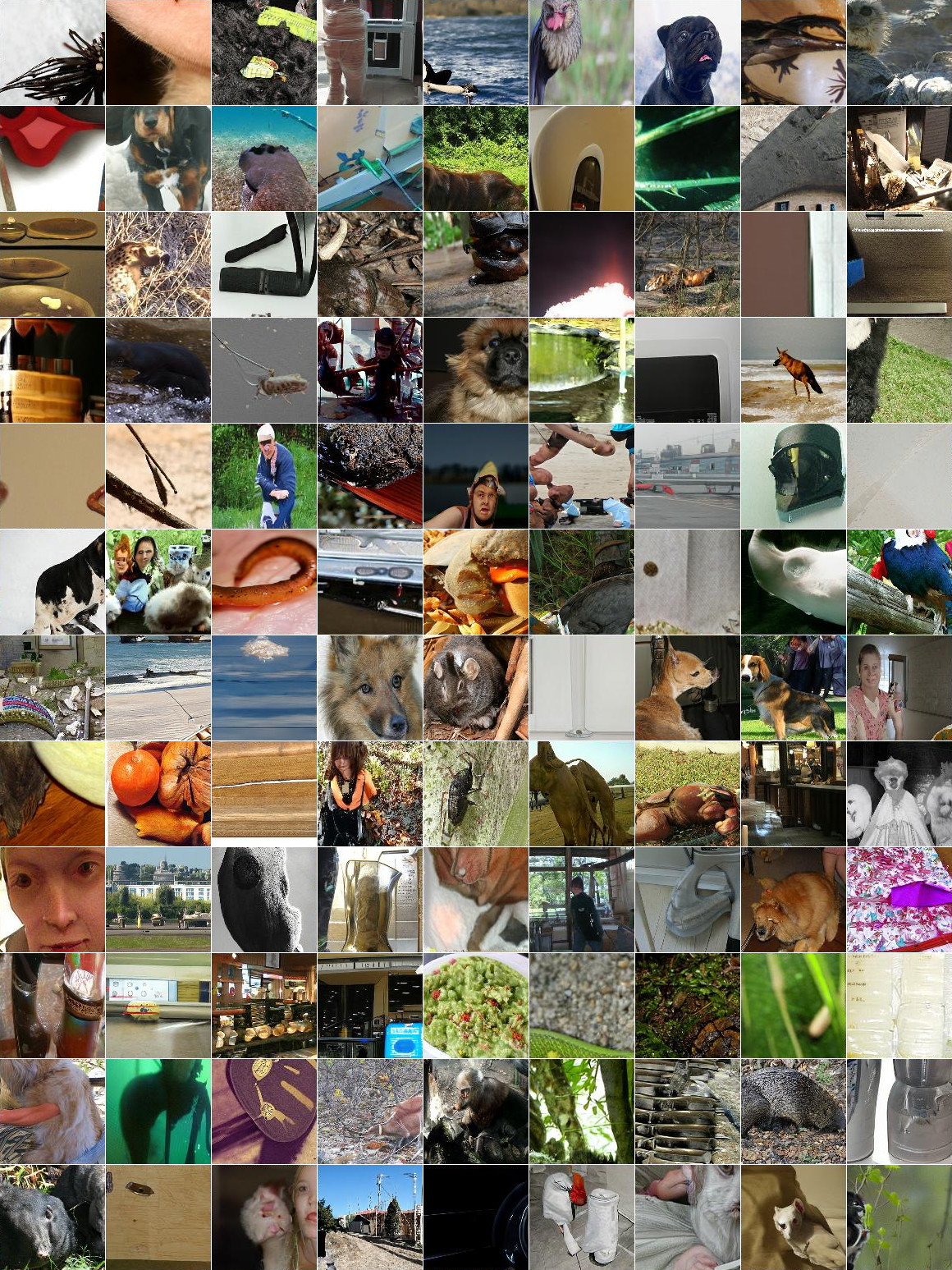}

 \caption{
  $128 \times 128$ samples $\hat{\x} \sim \gen(\z)$ from an unsupervised \method{} generator $\gen$, trained using the \textit{Base} method from Table~\ref{ablations}.
}
 \label{fig:samples_128}
\end{figure}

\begin{figure}
\centering

 \includegraphics[trim=0 1290 0 0,clip,width=1\textwidth]{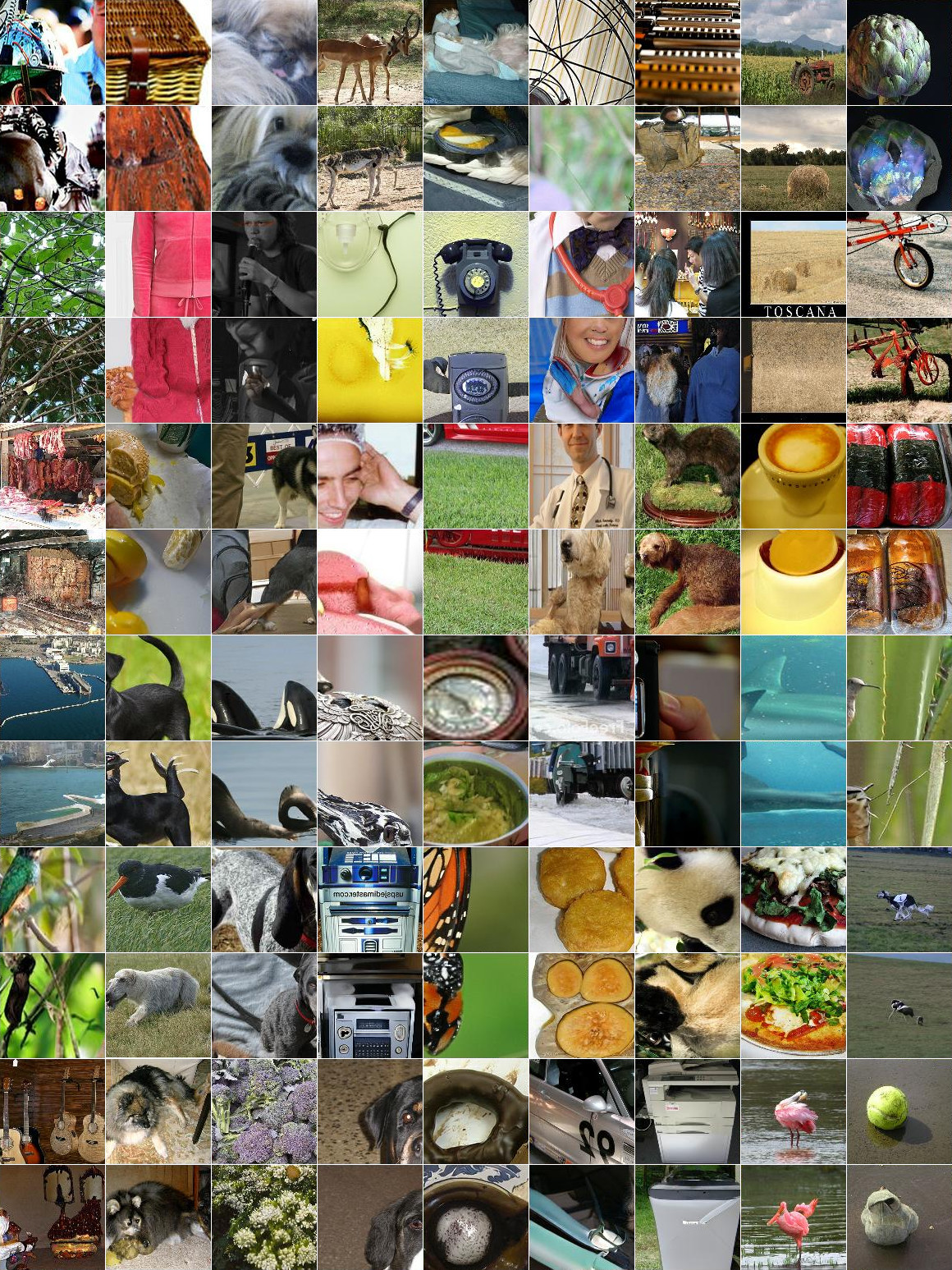}
 \\ \vspace{0.1cm}
 \includegraphics[trim=0 1032 0 258,clip,width=1\textwidth]{recons_base_6x9.jpg}
 \\ \vspace{0.1cm}
 \includegraphics[trim=0 774 0 516,clip,width=1\textwidth]{recons_base_6x9.jpg}
 \\ \vspace{0.1cm}
 \includegraphics[trim=0 516 0 774,clip,width=1\textwidth]{recons_base_6x9.jpg}
 \\ \vspace{0.1cm}
 \includegraphics[trim=0 258 0 1032,clip,width=1\textwidth]{recons_base_6x9.jpg}
 \\ \vspace{0.1cm}
 \includegraphics[trim=0 0 0 1290,clip,width=1\textwidth]{recons_base_6x9.jpg}
 \caption{
  $128 \times 128$ reconstructions from an unsupervised \method{} model, trained using the \textit{Base} method from Table~\ref{ablations}.
  The top rows of each pair are real data $\x \sim P_\x$, and bottom rows are generated reconstructions computed by $\gen(\enc(\x))$.
}
 \label{fig:more_recons_128}
\end{figure}

\begin{figure}
\centering

 \includegraphics[width=1\textwidth]{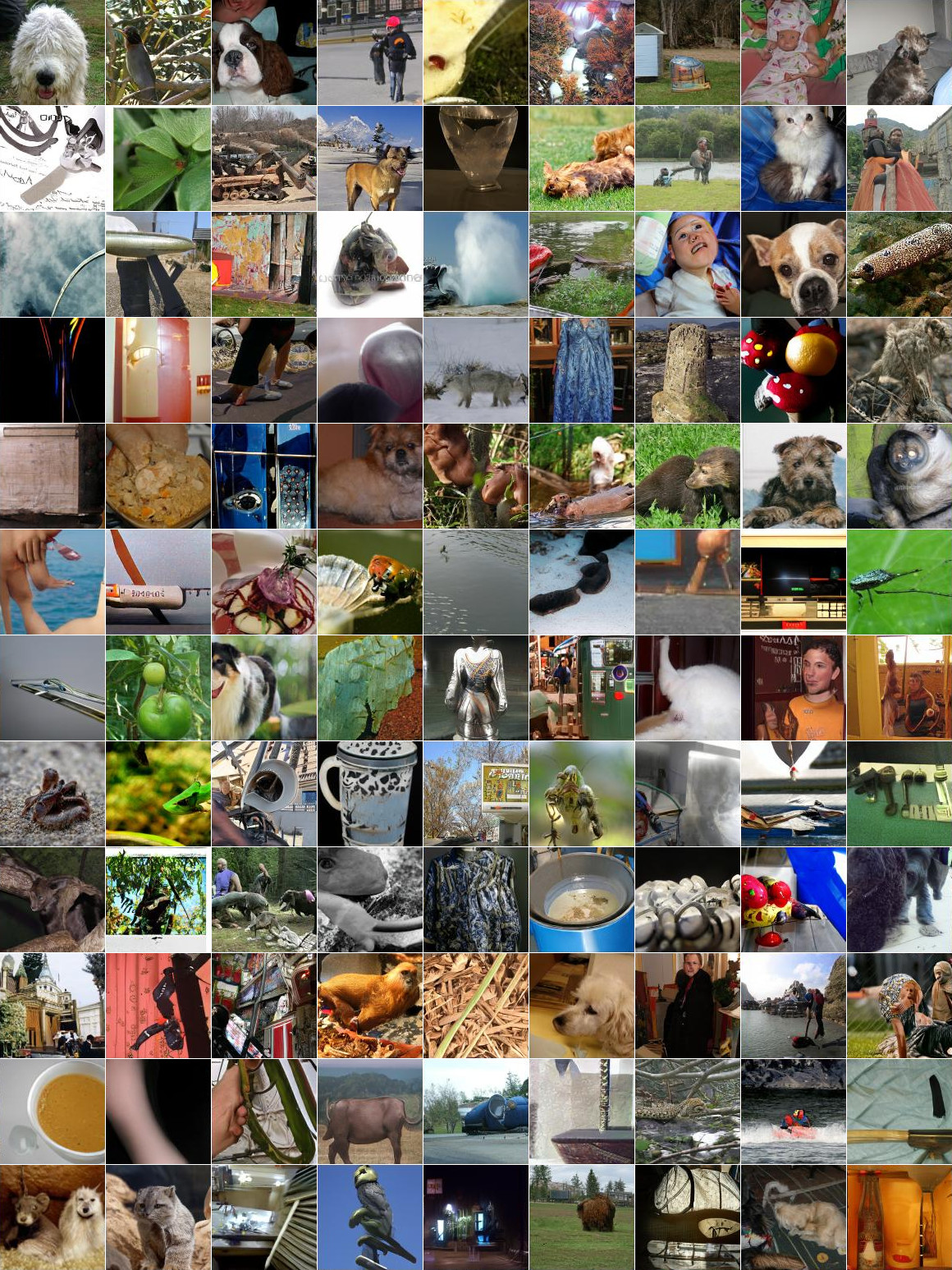}
 \caption{
  $128 \times 128$ samples $\hat{\x} \sim \gen(\z)$ from an unsupervised \method{} generator $\gen$, trained using the lighter augmentation from~\cite{zurichfewer} with generation results reported in Table~\ref{uncond_gen_results}.
 }
 \label{fig:samples_i128lightaug}
\end{figure}

\begin{figure}
\centering

 \includegraphics[trim=0 1290 0 0,clip,width=1\textwidth]{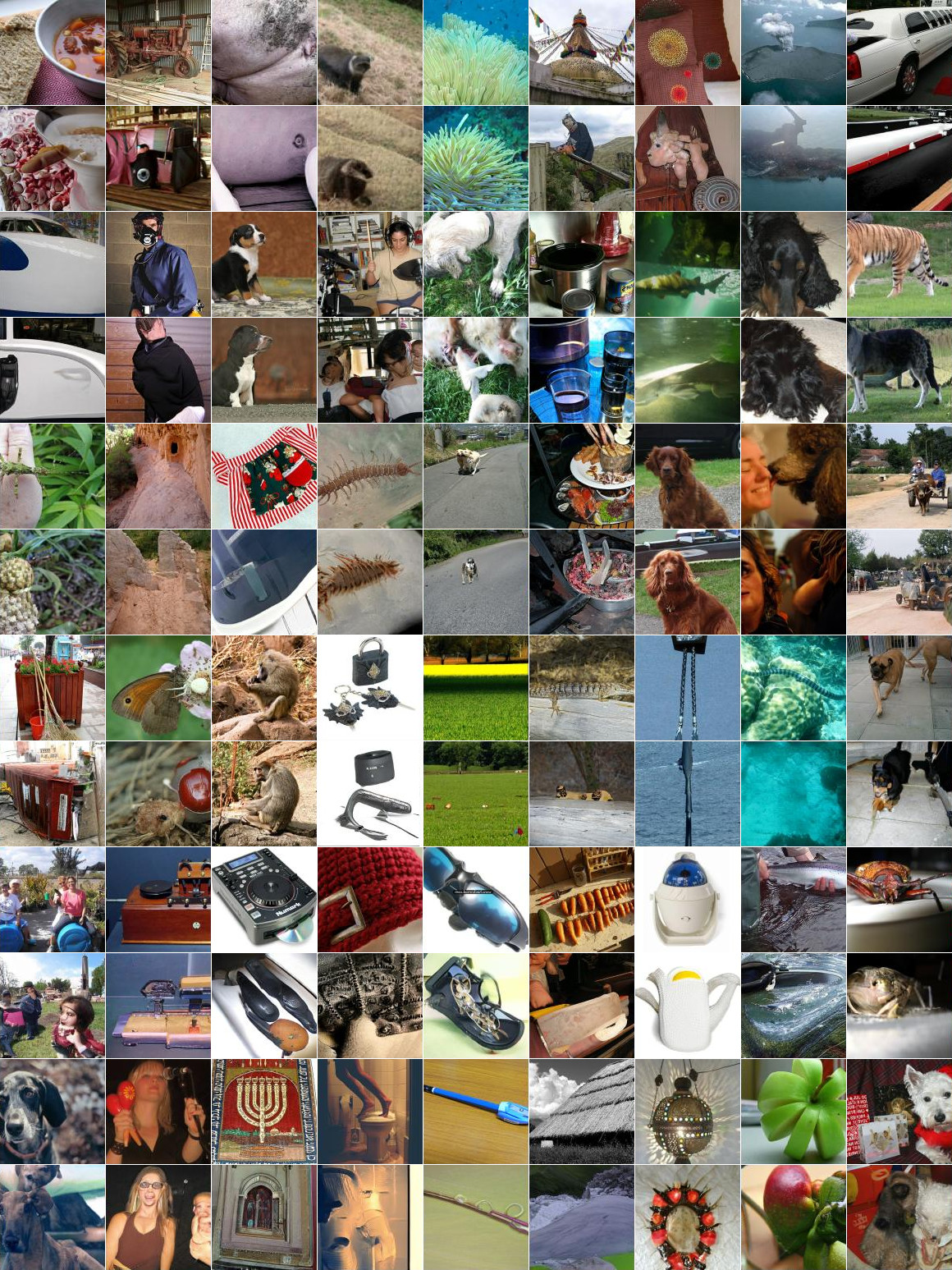}
 \\ \vspace{0.1cm}
 \includegraphics[trim=0 1032 0 258,clip,width=1\textwidth]{recons_i128df_6x9.jpg}
 \\ \vspace{0.1cm}
 \includegraphics[trim=0 774 0 516,clip,width=1\textwidth]{recons_i128df_6x9.jpg}
 \\ \vspace{0.1cm}
 \includegraphics[trim=0 516 0 774,clip,width=1\textwidth]{recons_i128df_6x9.jpg}
 \\ \vspace{0.1cm}
 \includegraphics[trim=0 258 0 1032,clip,width=1\textwidth]{recons_i128df_6x9.jpg}
 \\ \vspace{0.1cm}
 \includegraphics[trim=0 0 0 1290,clip,width=1\textwidth]{recons_i128df_6x9.jpg}
 \caption{
  $128 \times 128$ reconstructions from an unsupervised \method{} model, trained using the lighter augmentation from~\cite{zurichfewer} with generation results reported in Table~\ref{uncond_gen_results}.
  The top rows of each pair are real data $\x \sim P_\x$, and bottom rows are generated reconstructions computed by $\gen(\enc(\x))$.
}
 \label{fig:recons_i128lightaug}
\end{figure}

\begin{figure}
\centering

 \includegraphics[width=1\textwidth]{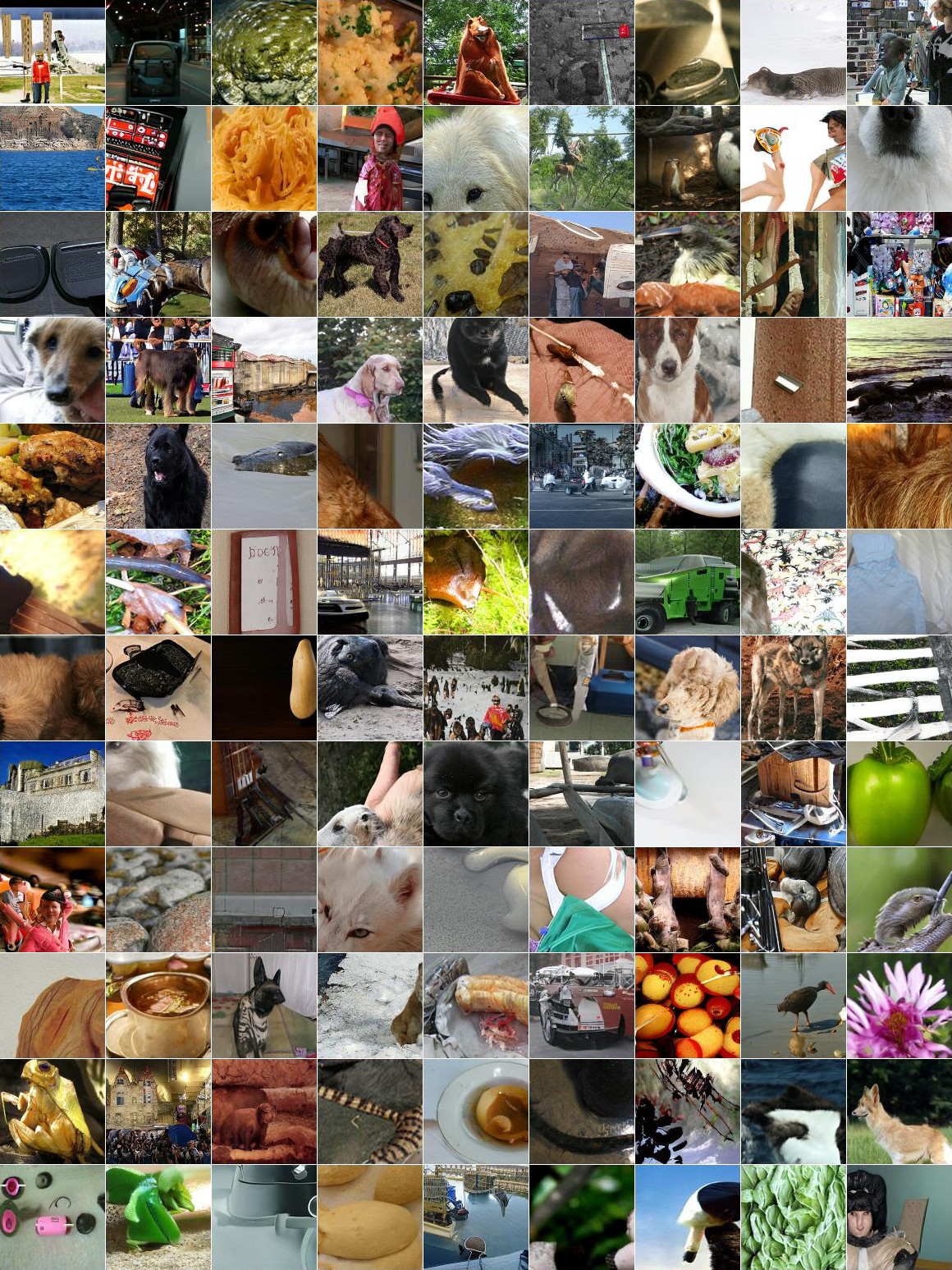}
 \caption{
  $128 \times 128$ samples $\hat{\x} \sim \gen(\z)$ from an unsupervised \method{} generator $\gen$, trained using the \textit{High Res $\enc$ (256)} configuration from Table~\ref{ablations}.}
 \label{fig:samples_highrese}
\end{figure}

\begin{figure}
\centering

 \includegraphics[trim=0 1542 0 0,clip,width=\textwidth]{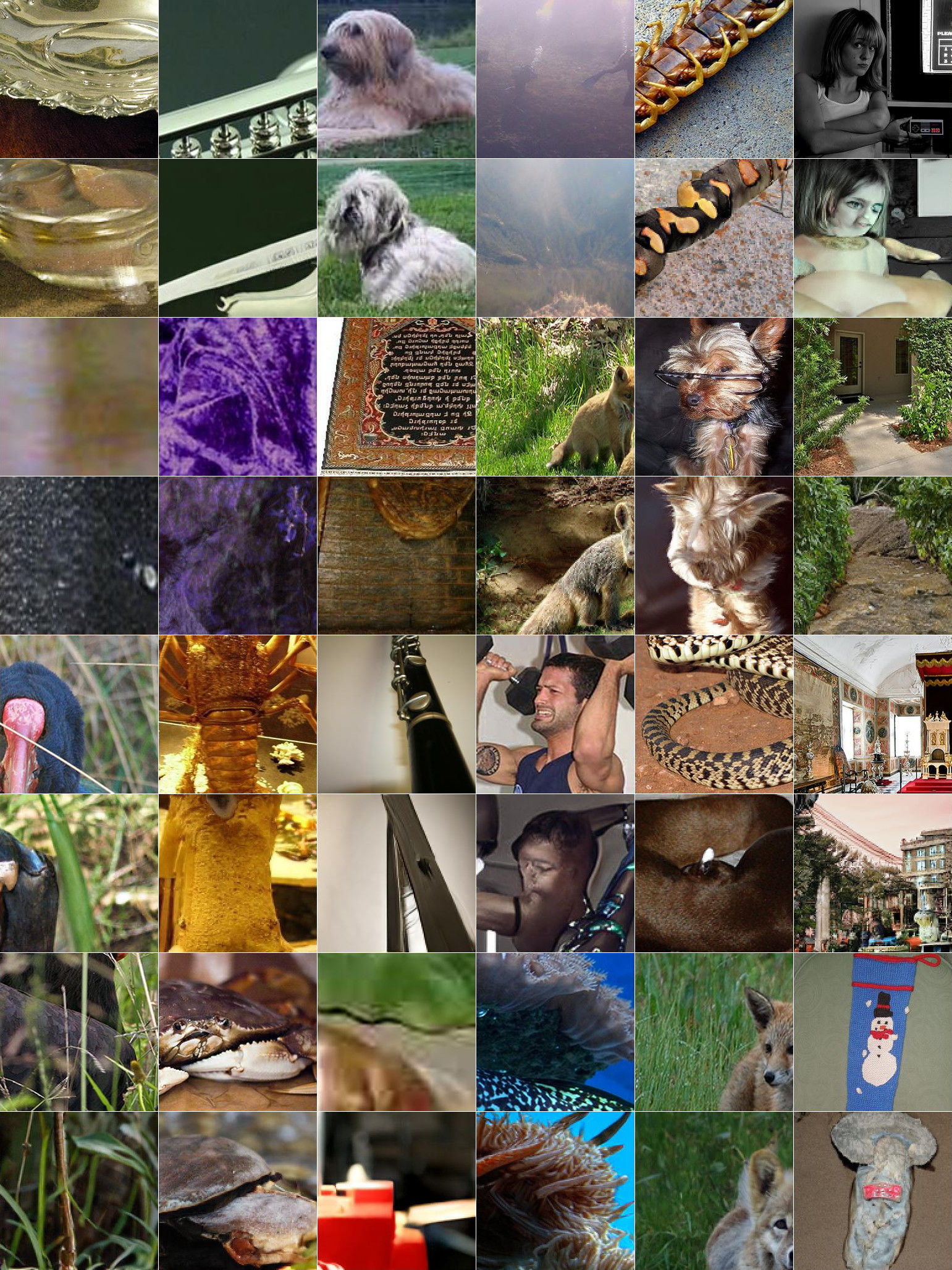}
 \\ \vspace{0.2cm}
 \includegraphics[trim=0 1028 0 514,clip,width=\textwidth]{recons_highresE_4x6_0_qual_80.jpg}
 \\ \vspace{0.2cm}
 \includegraphics[trim=0 514 0 1028,clip,width=\textwidth]{recons_highresE_4x6_0_qual_80.jpg}
 \\ \vspace{0.2cm}
 \includegraphics[trim=0 0 0 1542,clip,width=\textwidth]{recons_highresE_4x6_0_qual_80.jpg}

 \caption{
  $128 \times 128$ reconstructions of $256 \times 256$ encoder input images from an unsupervised \method{} model, trained using the \textit{High Res $\enc$ (256)} configuration from Table~\ref{ablations}.
  Reconstructions are upsampled from $128 \times 128$ to $256 \times 256$ for visualization.
  The top rows of each pair are real data $\x \sim P_\x$, and bottom rows are generated reconstructions computed by $\gen(\enc(\x))$.
}
 \label{fig:recons_highrese}
\end{figure}

\begin{figure}
\centering

 \includegraphics[width=1\textwidth]{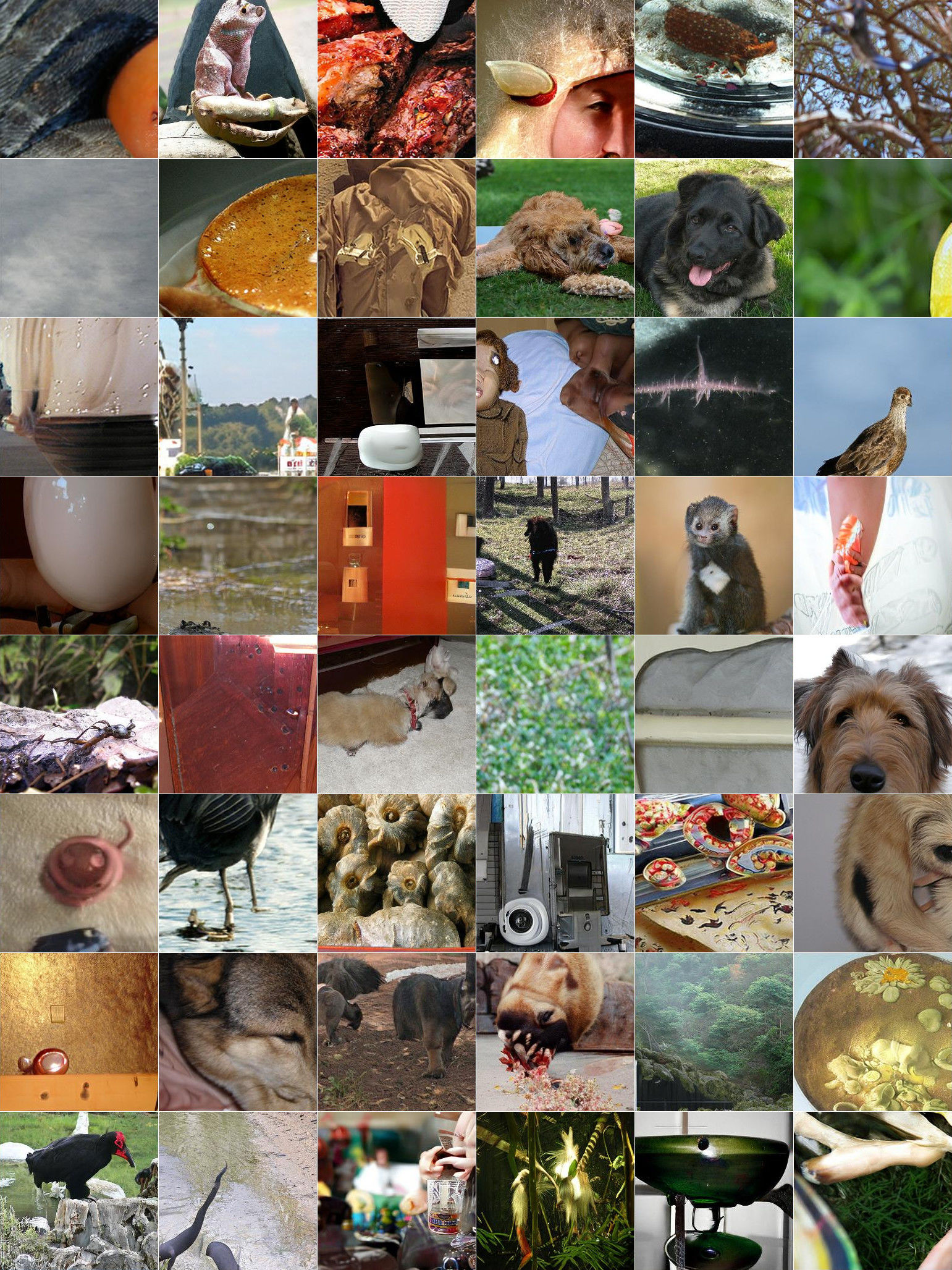}
 \caption{
  $256 \times 256$ samples $\hat{\x} \sim \gen(\z)$ from an unsupervised \method{} generator $\gen$, trained with a high-resolution $\enc$ and $\gen$ (\textit{High Res $\gen$ (256)} from Table~\ref{ablations}).
 }
 \label{fig:samples_256}
\end{figure}

\begin{figure}
\centering

 \includegraphics[trim=0 1542 0 0,clip,width=\textwidth]{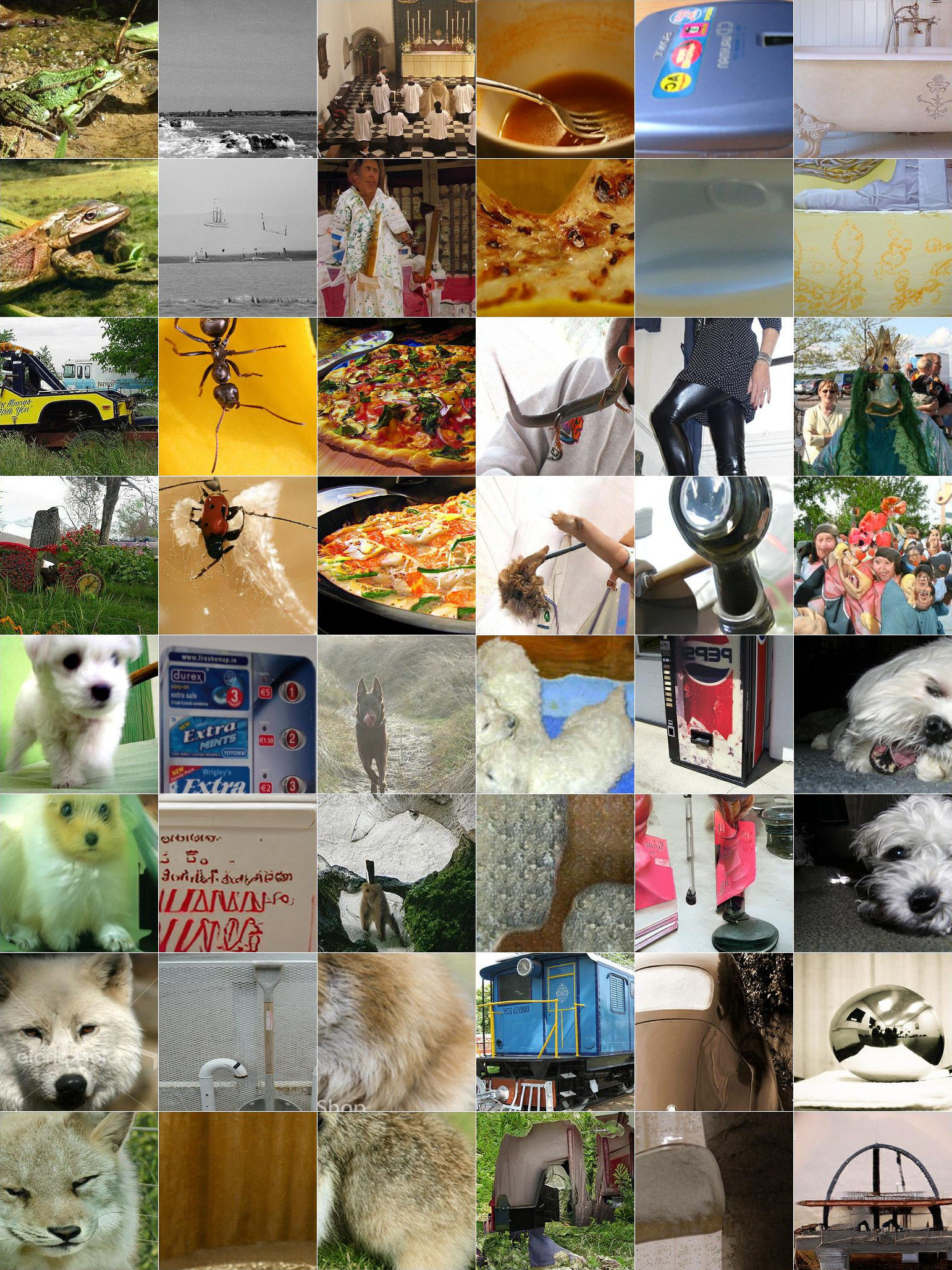}
 \\ \vspace{0.2cm}
 \includegraphics[trim=0 1028 0 514,clip,width=\textwidth]{recons_256_qual_80.jpg}
 \\ \vspace{0.2cm}
 \includegraphics[trim=0 514 0 1028,clip,width=\textwidth]{recons_256_qual_80.jpg}
 \\ \vspace{0.2cm}
 \includegraphics[trim=0 0 0 1542,clip,width=\textwidth]{recons_256_qual_80.jpg}
 \caption{
  $256 \times 256$ reconstructions from an unsupervised \method{} model, trained with a high-resolution $\enc$ and $\gen$ (\textit{High Res $\gen$ (256)} from Table~\ref{ablations}).
  The top rows of each pair are real data $\x \sim P_\x$, and bottom rows are generated reconstructions computed by $\gen(\enc(\x))$.
}
 \label{fig:recons_256}
\end{figure}

\setlength{\tabcolsep}{2pt}
\setlength{\tabcolsep}{6pt}

\newpage
\section{Nearest neighbors}
\label{appendix_neighbors}
\begin{table*}
\centering
\scalebox{1.}{
 \begin{tabular}{c|cccc}
 \toprule
                   & \multicolumn{4}{|c}{Top-1 / Top-5 Acc. (\%)} \\
  Metric & $k=1$ & $k=5$ & $k=25$ & $k=50$ \\
 \midrule
  $D_1$ & 38.09 / - & 41.28 / 58.56 & 43.32 / 65.12 & 42.73 / 66.22 \\
  $D_2$ & 35.68 / - & 38.61 / 55.59 & 40.65 / 62.23 & 40.15 / 63.42 \\
 \bottomrule
 \end{tabular}
}
 \caption{
  Accuracy of $k$ nearest neighbors classifiers in \method{} feature space on the ImageNet validation set.
  We report results under the normalized $\ell_1$ distance $D_1$
  as well as the normalized $\ell_2$ (cosine) distance $D_2$.
 }
 \label{neighbors_results}
\end{table*}

\begin{samepage}
In this Appendix we consider an alternative way of evaluating representations –- by means of $k$ nearest neighbors classification,
which does not involve learning any parameters during evaluation and is even simpler than learning a linear classifier as done in Section~\ref{sec:eval}.
For all results in this section, we use the outputs of the global average pooling layer (a flat 8192D feature) of our best performing model,
\textit{RevNet $\times 4$, $\uparrow \enc$ LR}.
We do not do any data augmentation for either the training or validation sets:
we simply crop each image at the center of its larger axis and resize to $256\times256$.

We use a normalized $\ell_1$ or $\ell_2$ distance metric as our nearest neighbors criterion, defined as $
D_p(a,b)
= \left|\left|\frac{a}{||a||_p} - \frac{b}{||b||_p}\right|\right|_p$, for $p \in \{1, 2\}$.
($D_2$ corresponds to cosine distance.)
For label predictions with multiple neighbors ($k>1$),
we use a simple counting scheme: the label with the most votes is selected as the prediction.
Ties (multiple labels with the same number of votes) are broken by $k=1$
 nearest neighbor classification among the data with the tied labels.

\end{samepage}

\paragraph{Quantitative results.}
In Table~\ref{neighbors_results} we present $k$ nearest neighbors classification results for $k \in \{1, 5, 25, 50\}$.
Across all $k$, the $\ell_1$-based metric $D_1$ outperforms $D_2$, and the remainder of our discussion refers to the $D_1$ results.
With just a single neighbor ($k=1$) we achieve a top-1 accuracy around 38\%.
Top-1 accuracy reaches 43\% with $k=25$, dropping off slightly at $k=50$ as votes from more distant neighbors are added.

\begin{figure}
\centering
 \includegraphics[trim=0 3084 771 0,clip,height=0.85\textheight]{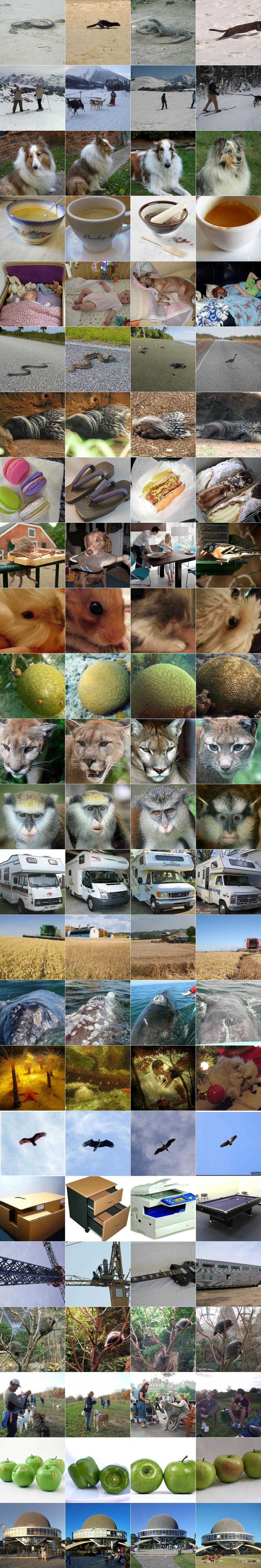}
 \includegraphics[trim=257 3084 0 0,clip,height=0.85\textheight]{neighbors_val_1_24_res256_k3_qual_50.jpg}
 \hspace{0.5cm}
 \includegraphics[trim=0 0 771 3084,clip,height=0.85\textheight]{neighbors_val_1_24_res256_k3_qual_50.jpg}
 \includegraphics[trim=257 0 0 3084,clip,height=0.85\textheight]{neighbors_val_1_24_res256_k3_qual_50.jpg}
 \caption{
  Nearest neighbors in \method{} $\enc$ feature space, from our best performing model (\textit{RevNet $\times 4$, $\uparrow \enc$ LR}).
  In each row, the first (left) column is a query image, and the remaining columns are its three nearest neighbors from the training set (the leftmost being the nearest, next being the second nearest, etc.).
  The query images above are the first 24 images in the ImageNet validation set.
}
 \label{fig:neighbors1}
\end{figure}

\paragraph{Qualitative results.}
Figure~\ref{fig:neighbors1} shows sample nearest neighbors in the ImageNet training set for query images in the validation set.
Despite being fully unsupervised, the neighbors in many cases match the query image in terms of high-level semantic content such as the category of the object of interest,
demonstrating \method{}'s ability to capture high-level attributes of the data in its unsupervised representations.
Where applicable, the object's pose and position in the image appears to be important as well -- for example, the nearest neighbors of the RV (row 2, column 2) are all RVs facing roughly the same direction.
In other cases, the nearest neighbors appear to be selected primarily based on the background or color scheme.

\paragraph{Discussion.}
While our quantitative $k$ nearest neighbors classification results are far from the state of the art for ImageNet classification
and significantly below the linear classifier-based results reported in Table~\ref{final_results},
note that in this setup, no supervised learning of model parameters from labels occurs at any point:
labels are predicted purely based on distance in a feature space
learned from \method{} training on image pixels alone.
We believe this makes nearest neighbors classification an interesting additional benchmark
for future approaches to unsupervised representation learning.

\newpage
\section{Learning curves}
\label{appendix_curves}
\begin{figure}
\centering
 \includegraphics[width=1.00\textwidth]{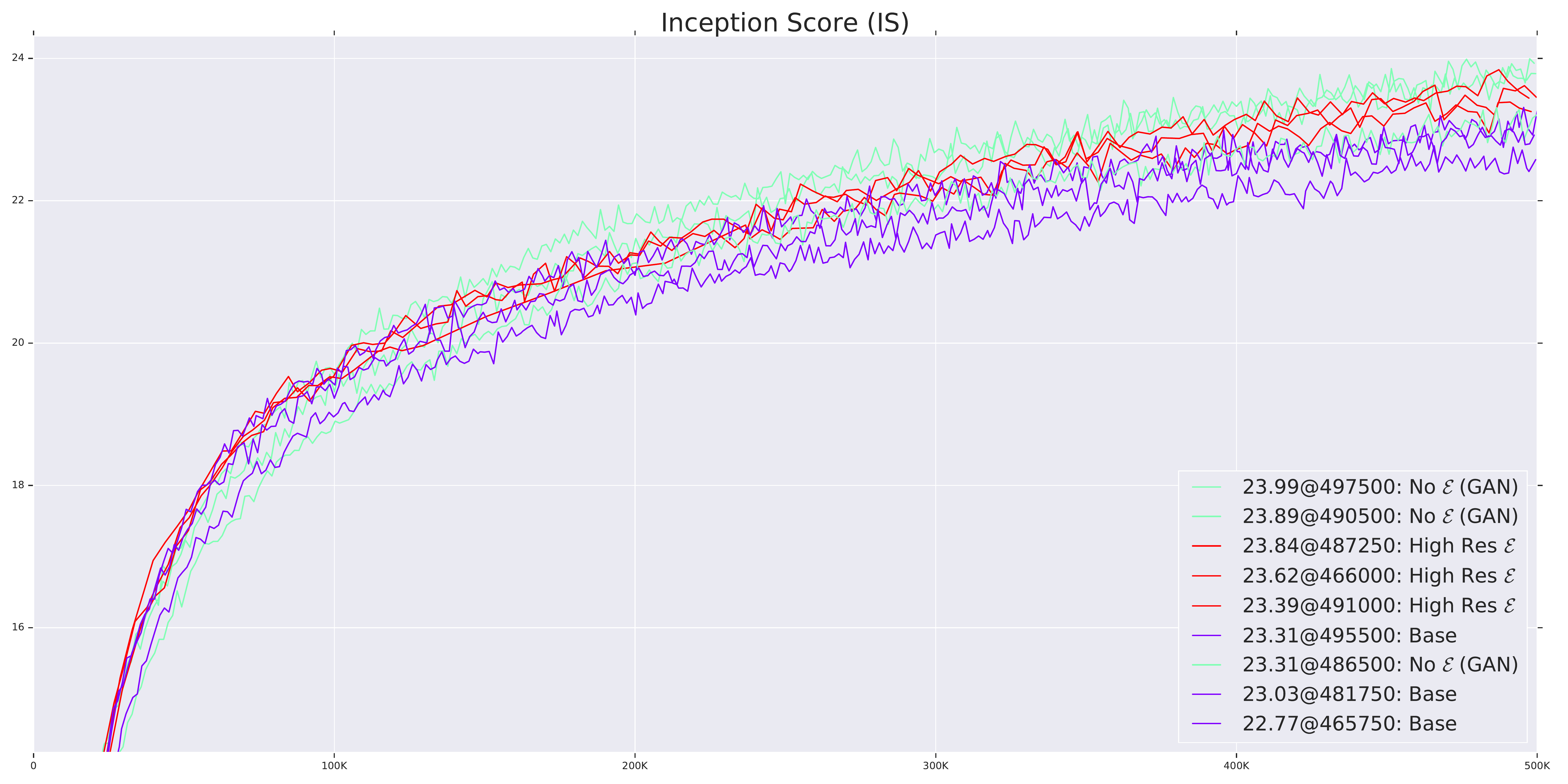}
 \includegraphics[width=1.00\textwidth]{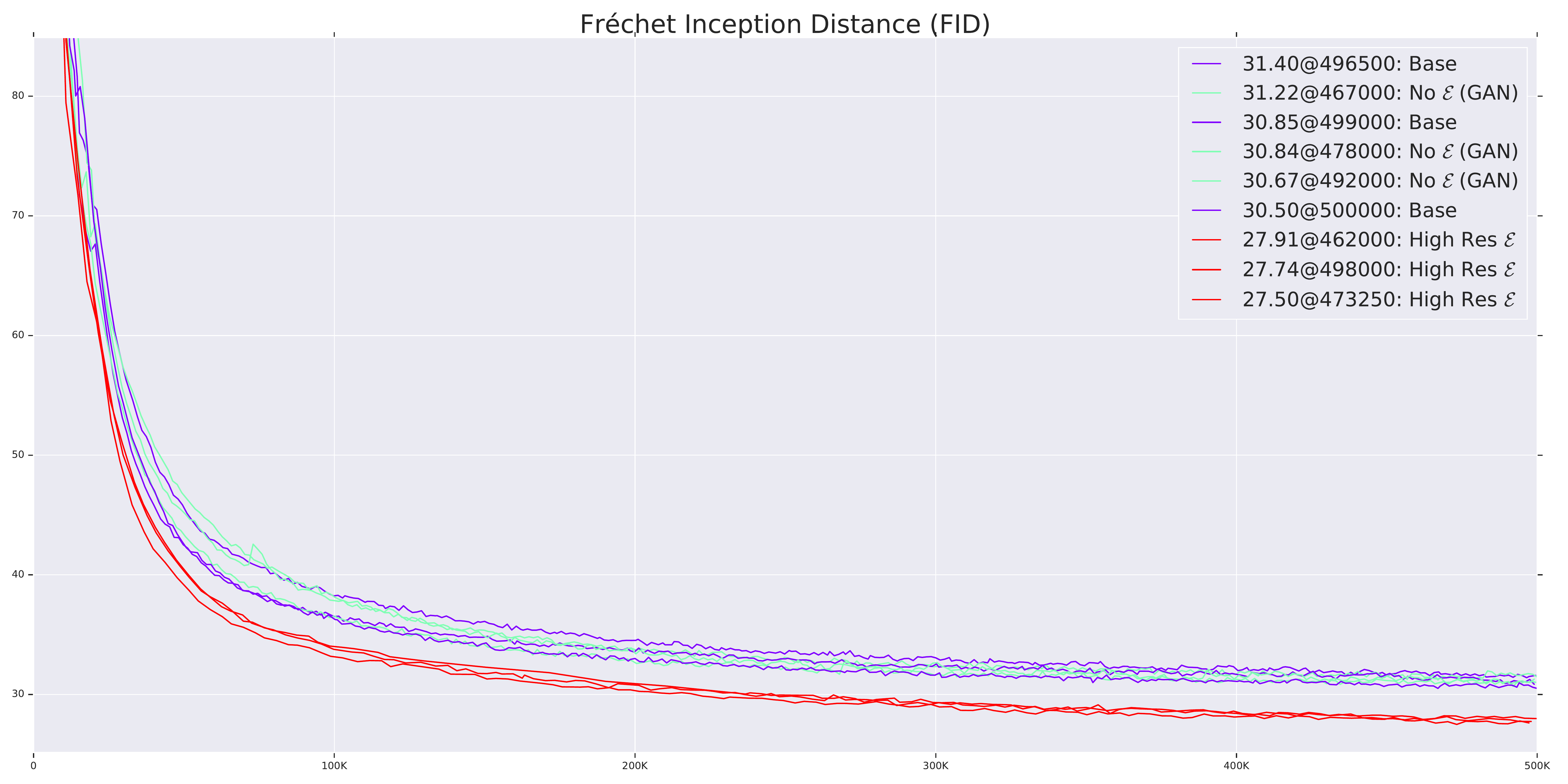}
 \caption{
   Image generation learning curves for several of the ablations in Section~\ref{sec:eval},
   including a comparison of \method{} to standard GAN.
   Legend entries correspond to the following rows in Table~\ref{ablations}:
\textit{Base},
\textit{No $\enc$ (GAN)}, and
\textit{High Res $\enc$ (256)}.
}
 \label{fig:curves_gen}
\end{figure}

\begin{figure}
\centering
 \includegraphics[width=1.00\textwidth]{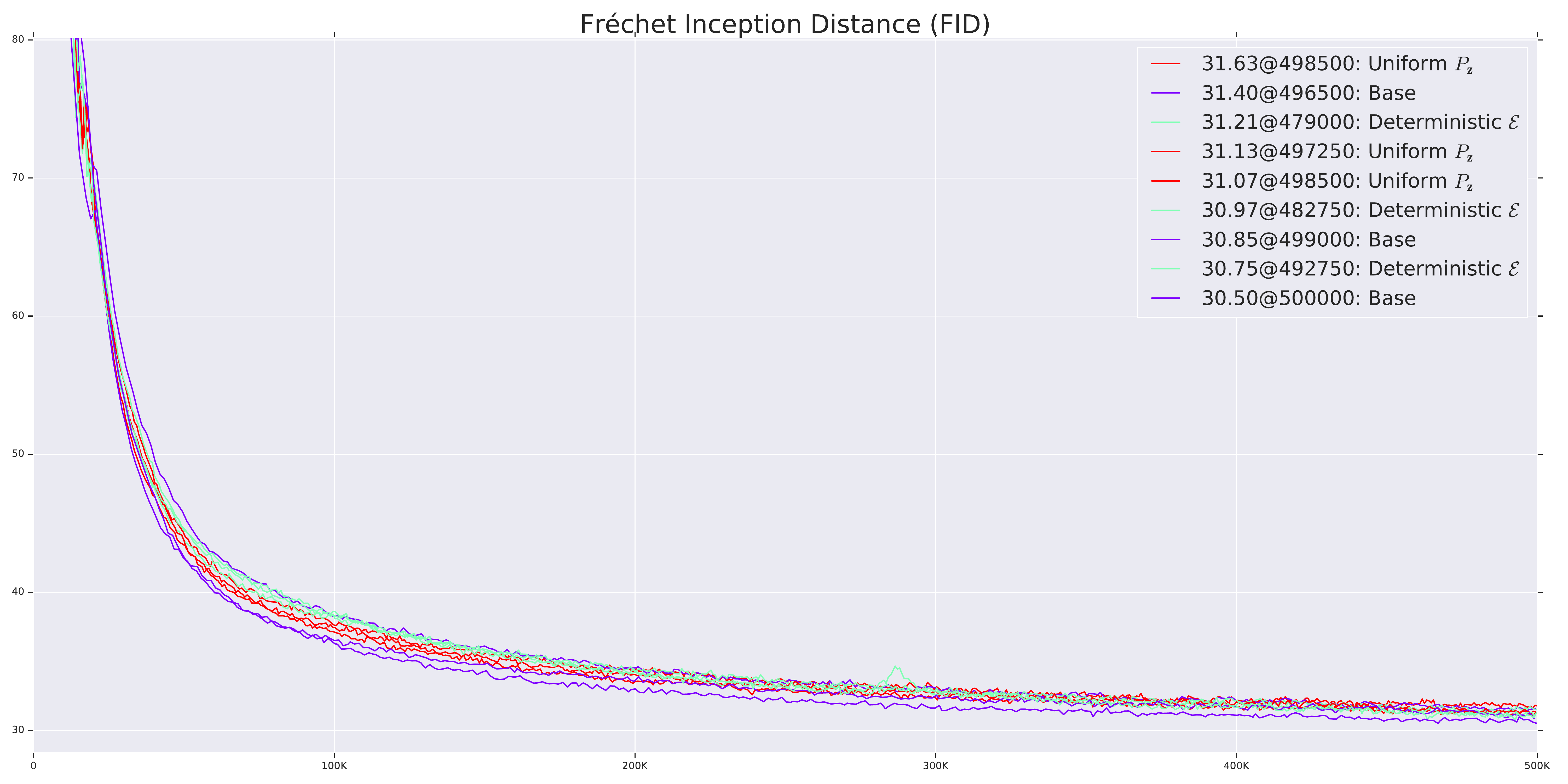}
 \includegraphics[width=1.00\textwidth]{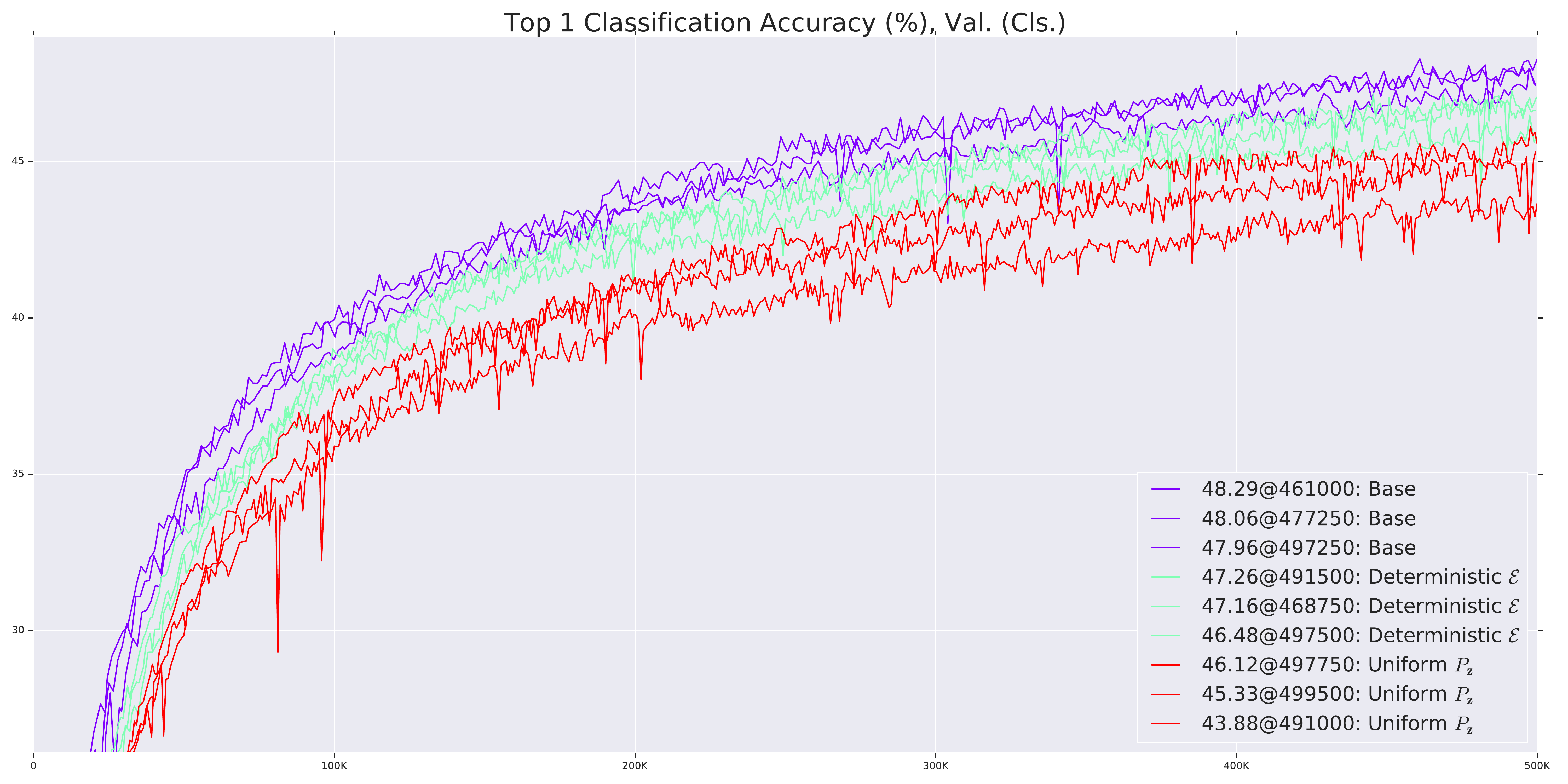}
 \includegraphics[width=1.00\textwidth]{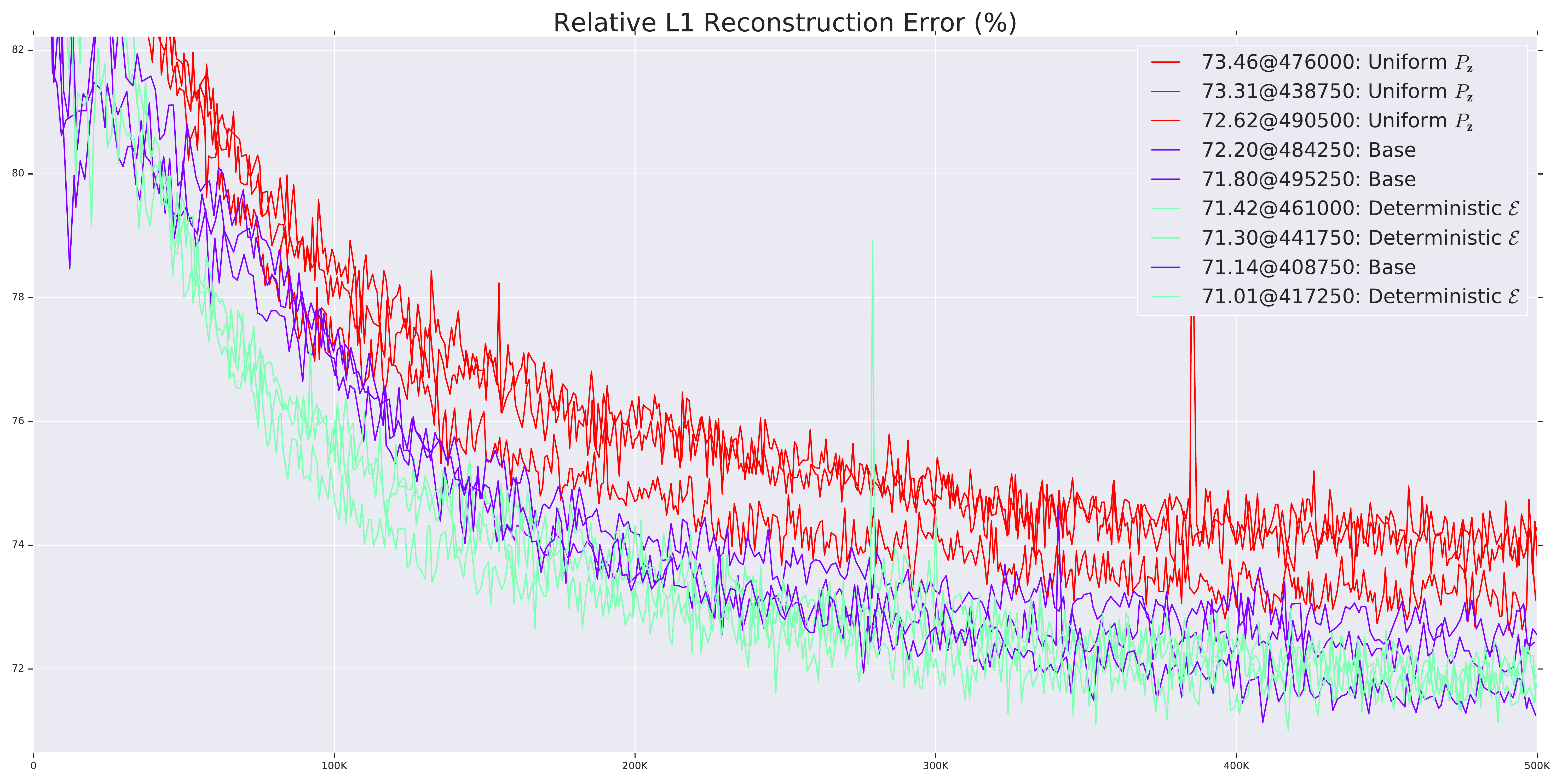}
 \caption{
   Image generation and representation learning curves for the latent space variations explored in Section~\ref{sec:eval}.
   Legend entries correspond to the following rows in Table~\ref{ablations}:
\textit{Base},
\textit{Deterministic $\enc$}, and
\textit{Uniform $P_\z$}.
}
 \label{fig:curves_latent}
\end{figure}

\begin{figure}
\centering
 \includegraphics[width=1.00\textwidth]{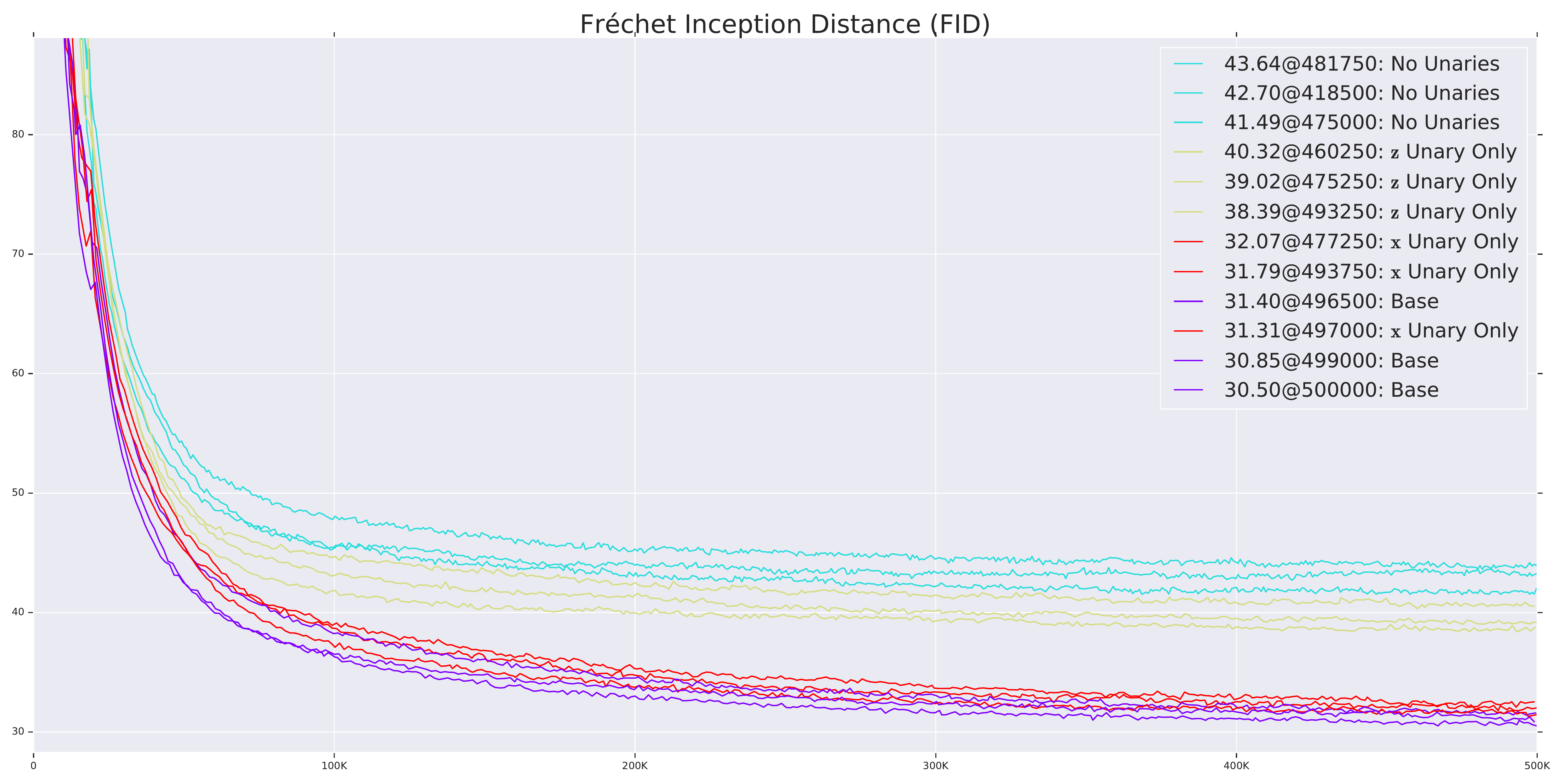}
 \includegraphics[width=1.00\textwidth]{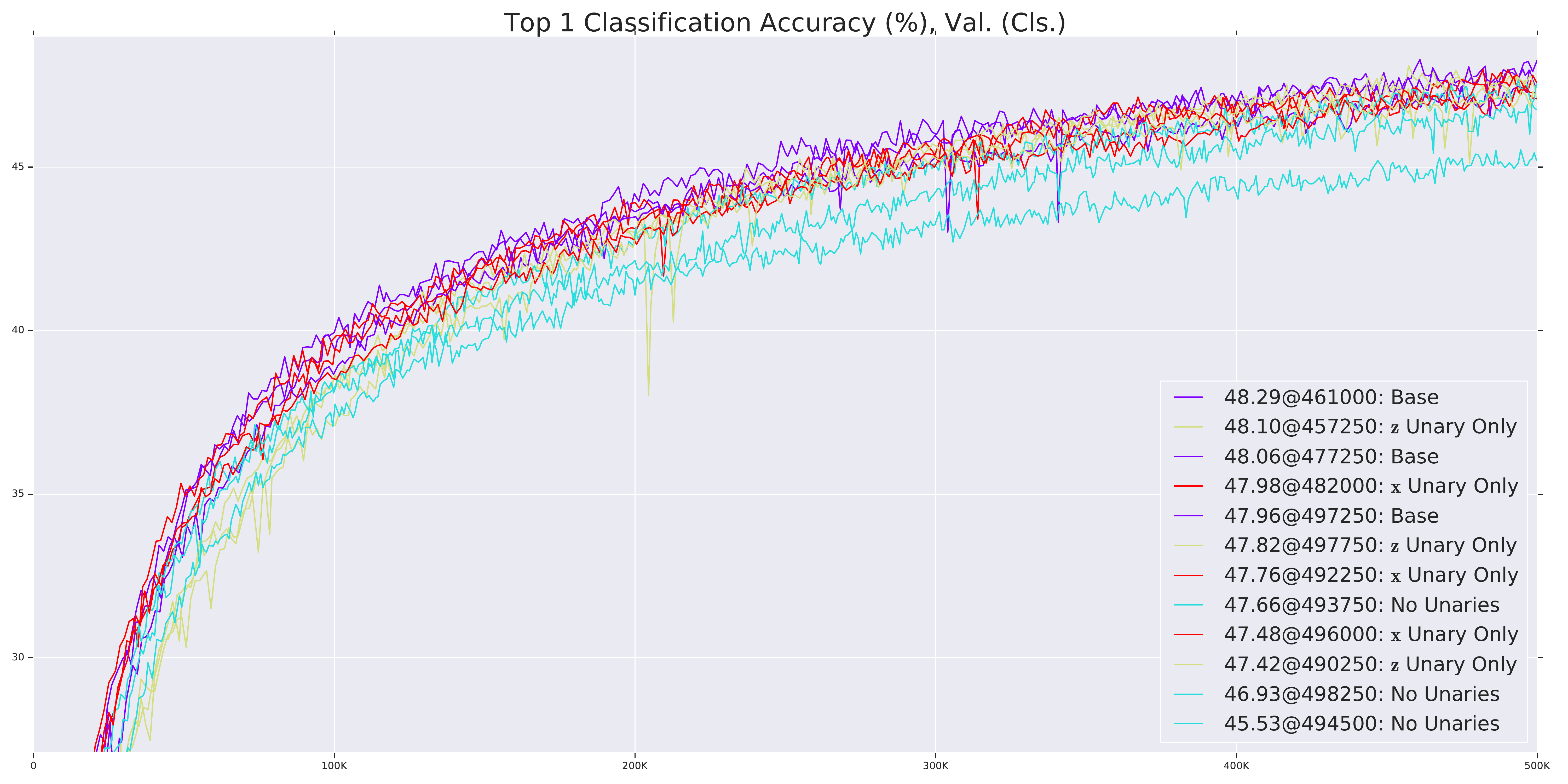}
 \includegraphics[width=1.00\textwidth]{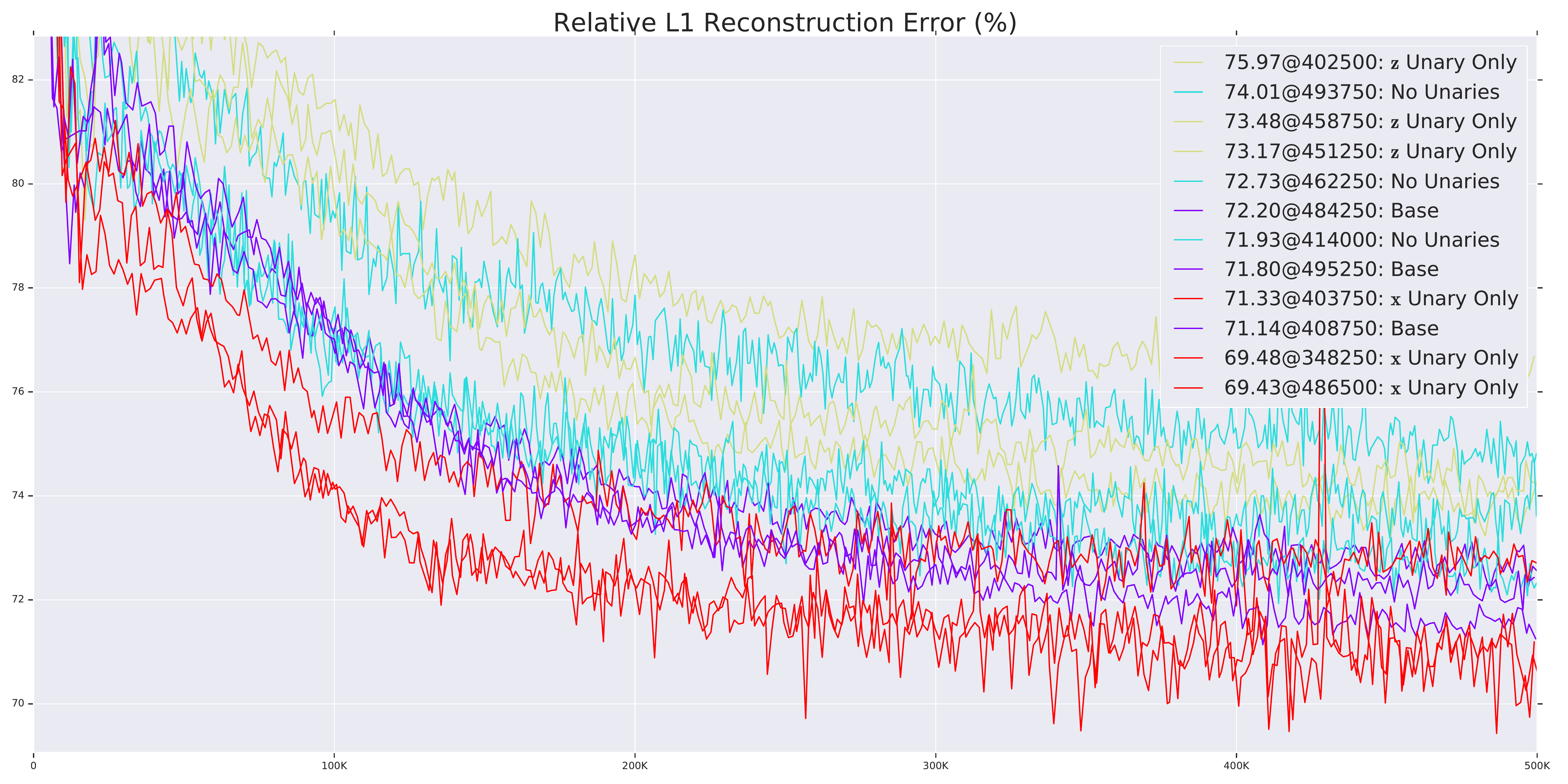}
 \caption{
   Image generation and representation learning curves for the unary loss component variations explored in Section~\ref{sec:eval}.
   Legend entries correspond to the following rows in Table~\ref{ablations}:
\textit{Base},
\textit{$\x$ Unary Only},
\textit{$\z$ Unary Only}, and
\textit{No Unaries (BiGAN)}.
}
 \label{fig:curves_unary}
\end{figure}

\begin{figure}
\centering
 \includegraphics[width=1.00\textwidth]{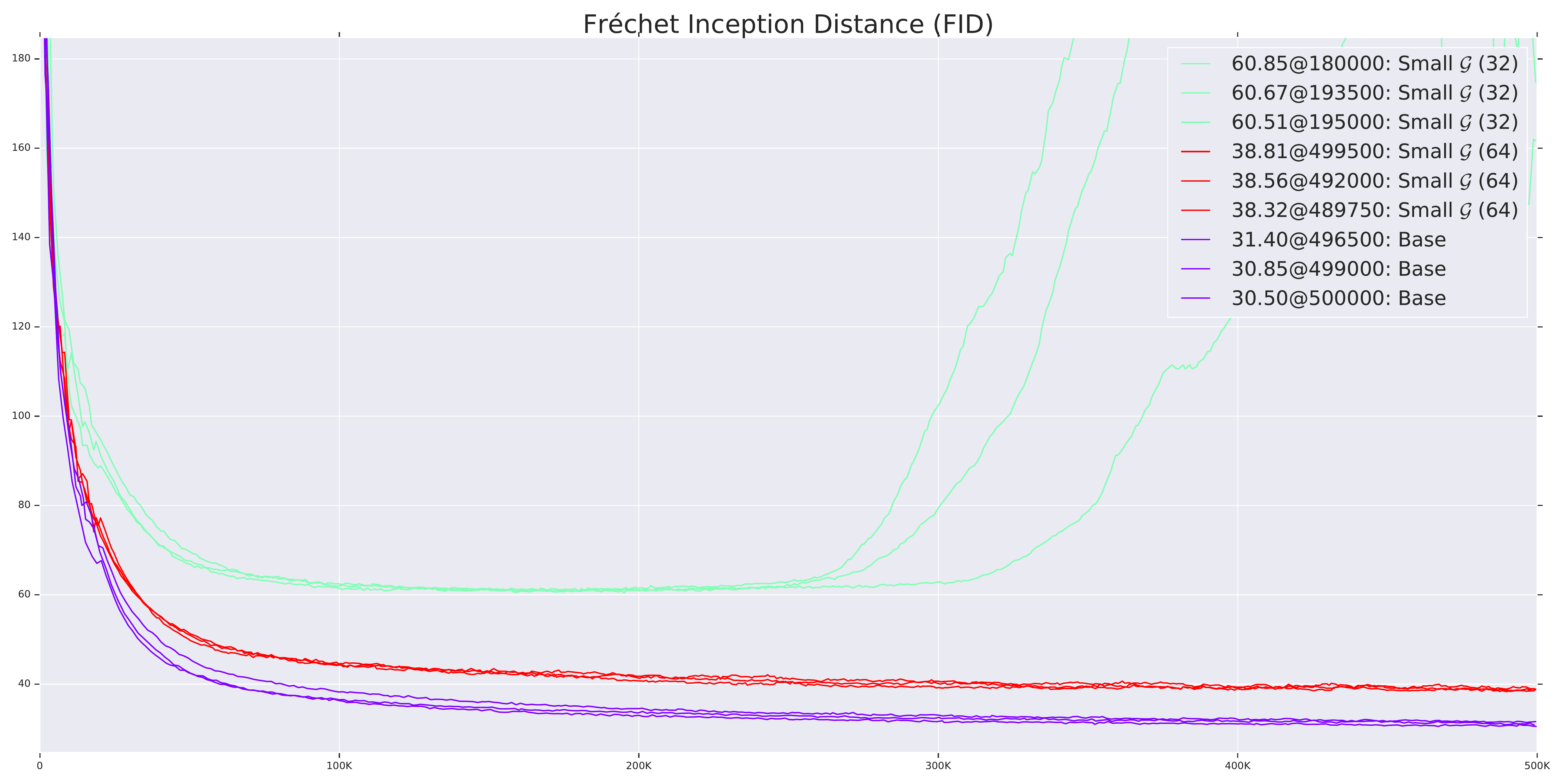}
 \includegraphics[width=1.00\textwidth]{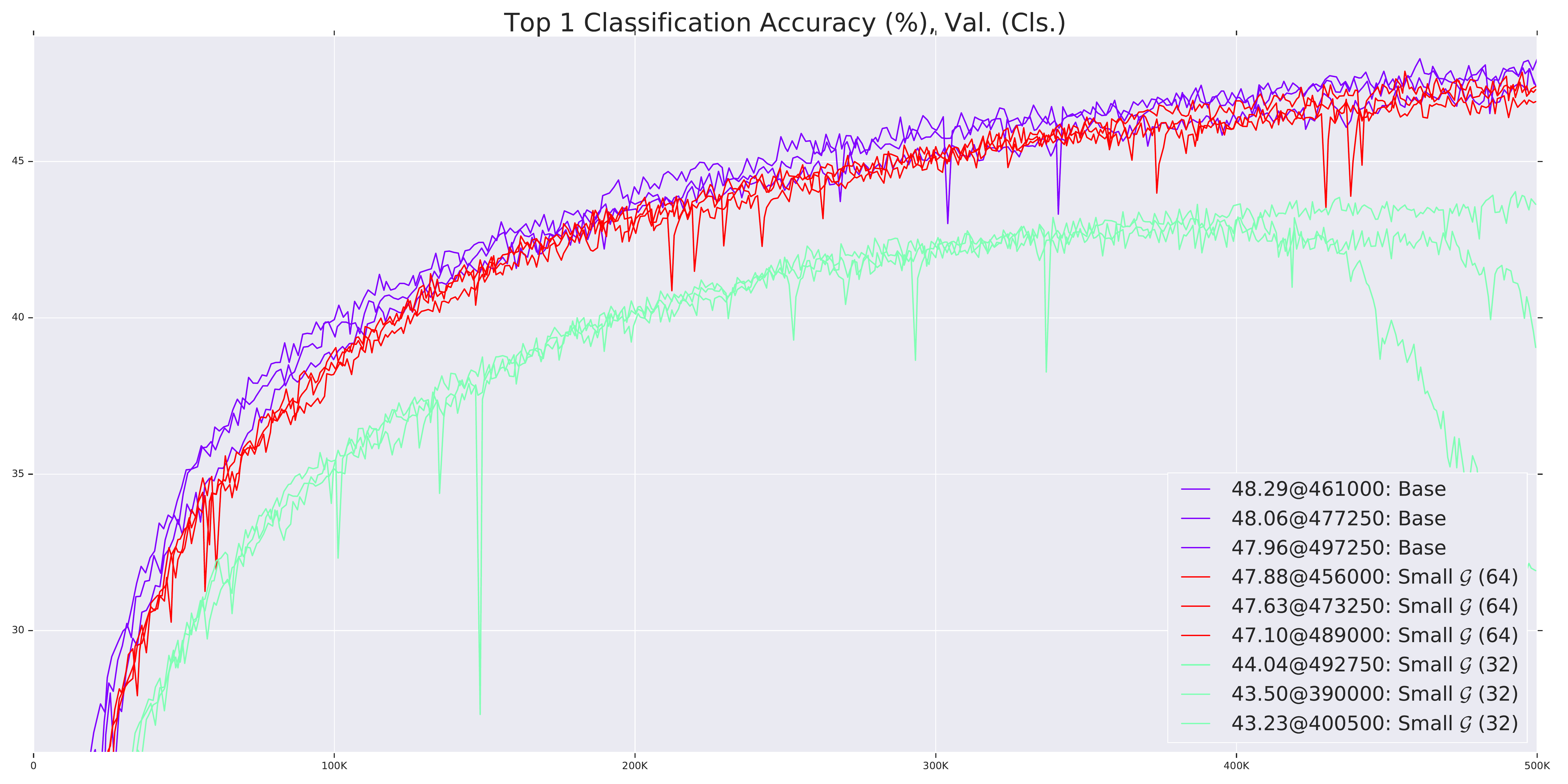}
 \includegraphics[width=1.00\textwidth]{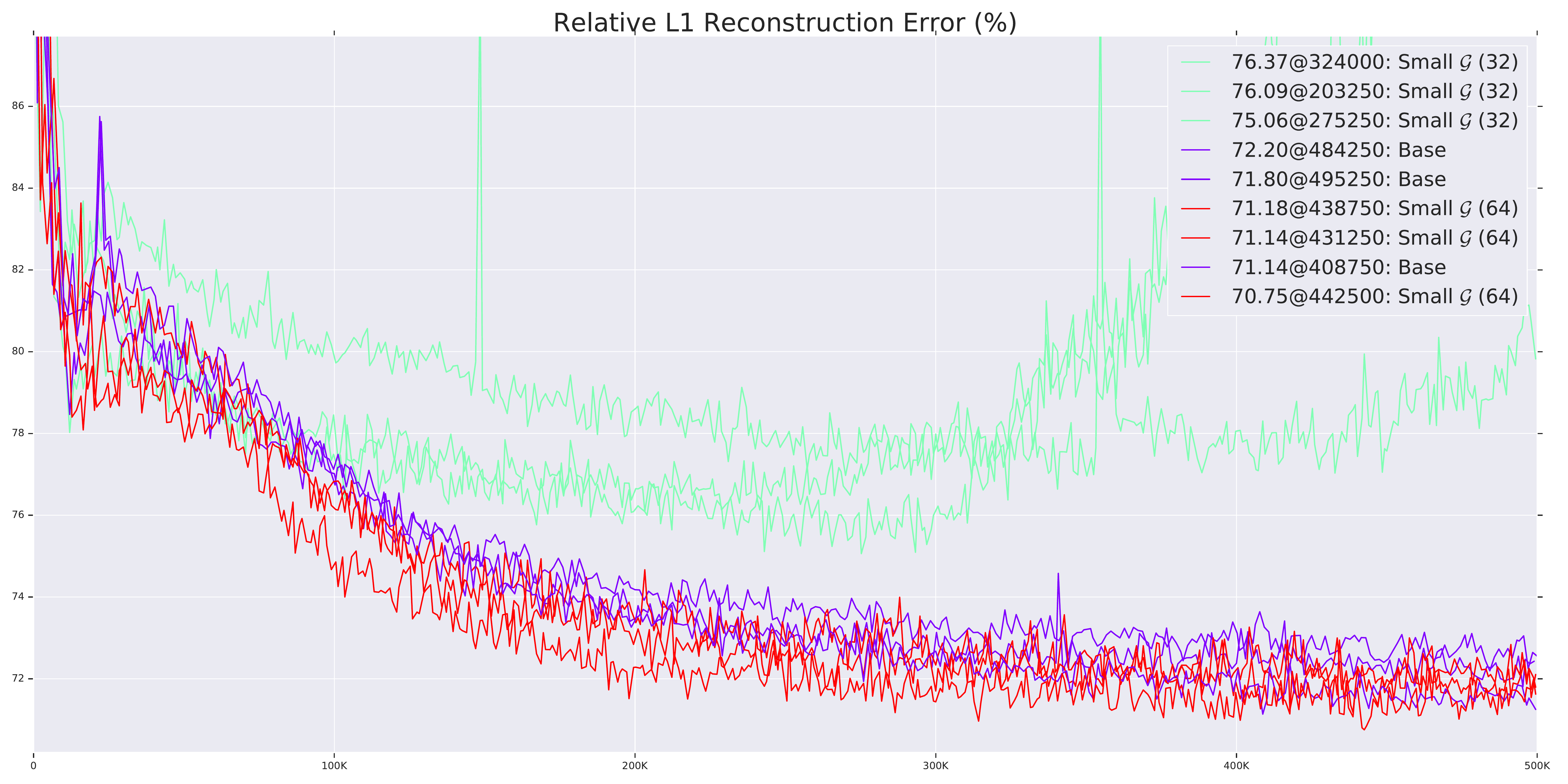}
 \caption{
   Image generation and representation learning curves for the $\gen$ size variations explored in Section~\ref{sec:eval}.
   Legend entries correspond to the following rows in Table~\ref{ablations}:
\textit{Base},
\textit{Small $\gen$ (32)}, and
\textit{Small $\gen$ (64)}.
}
 \label{fig:curves_small_g}
\end{figure}

\begin{figure}
\centering
 \includegraphics[width=1.00\textwidth]{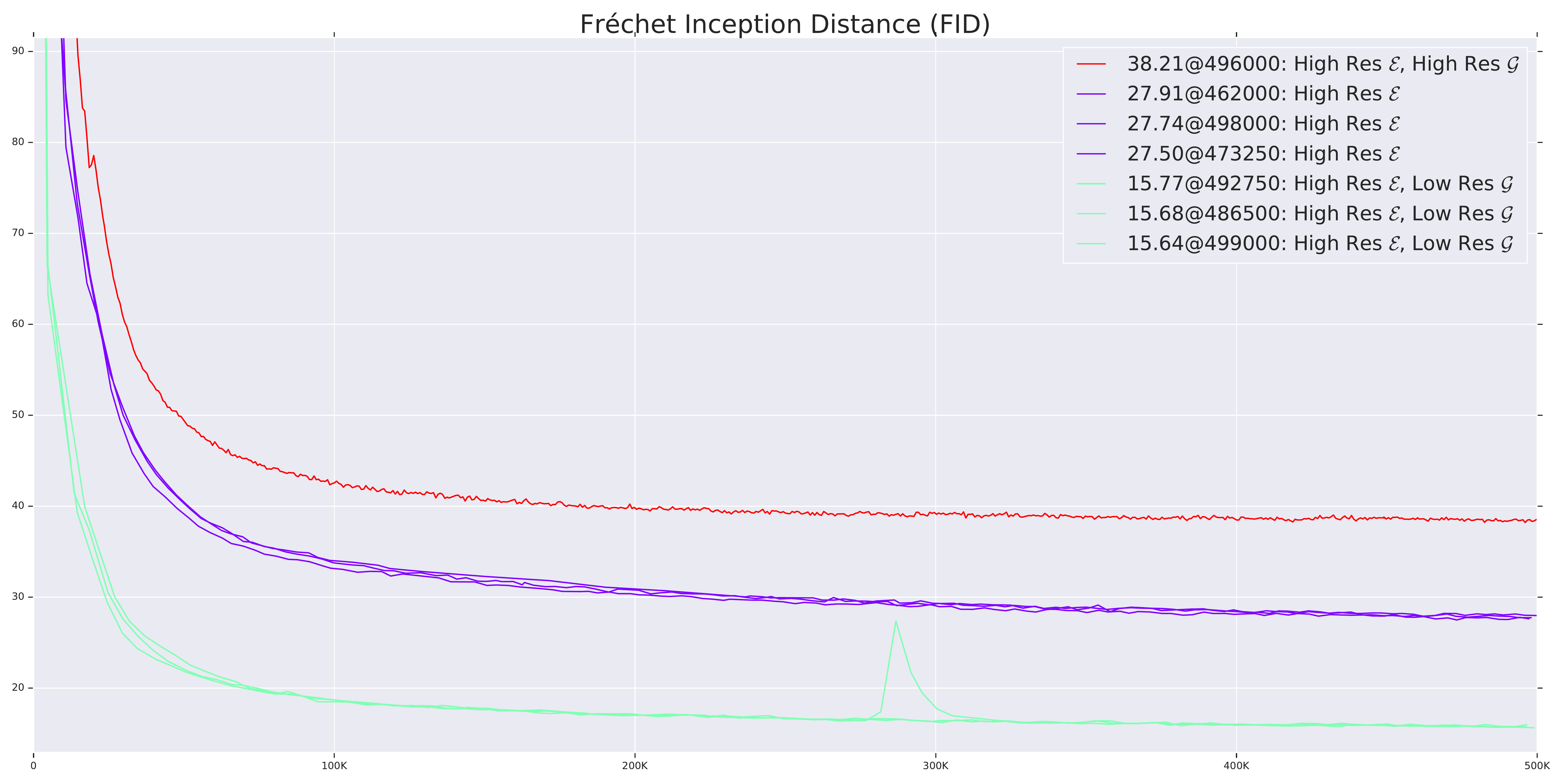}
 \includegraphics[width=1.00\textwidth]{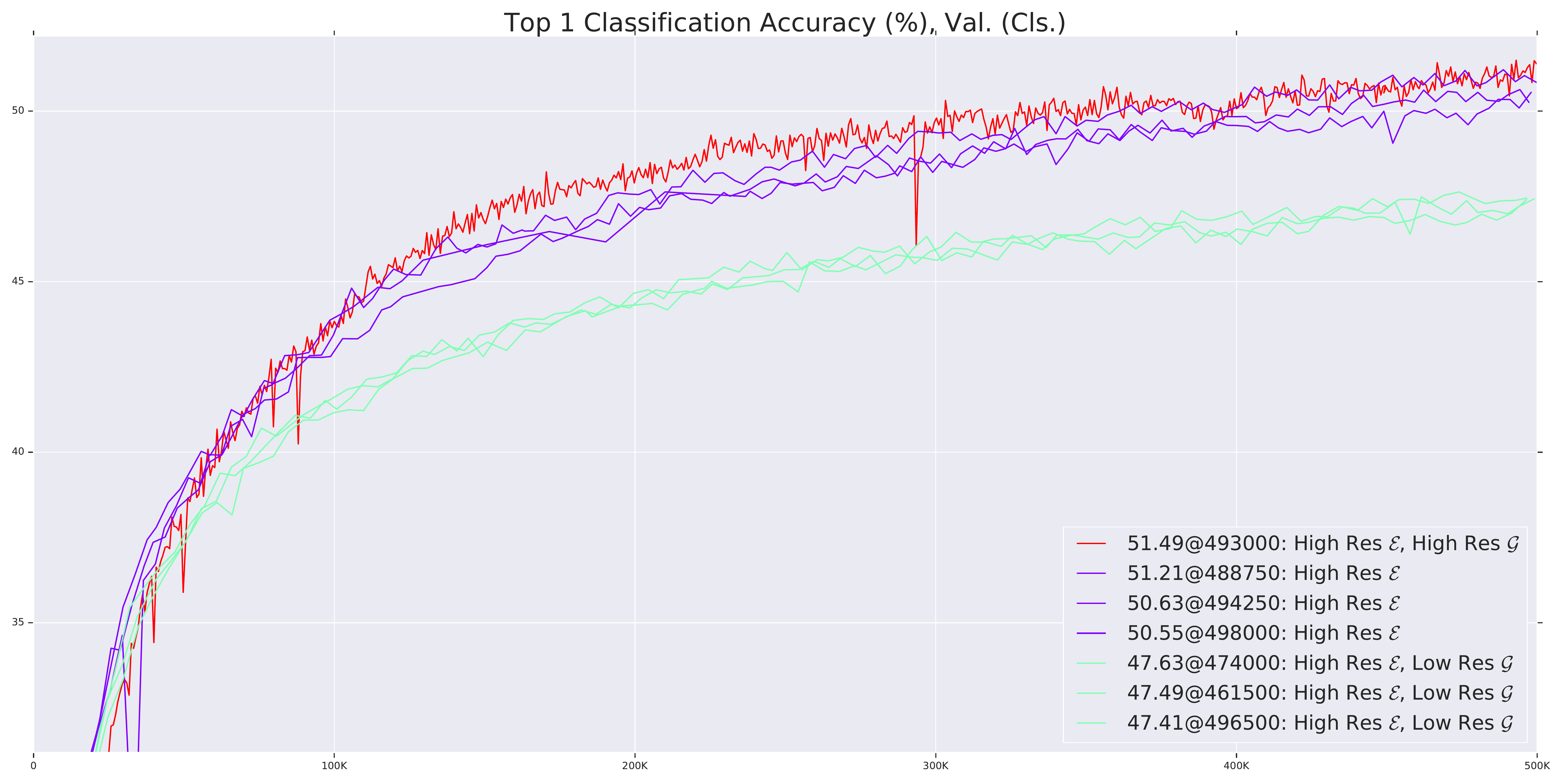}
 \includegraphics[width=1.00\textwidth]{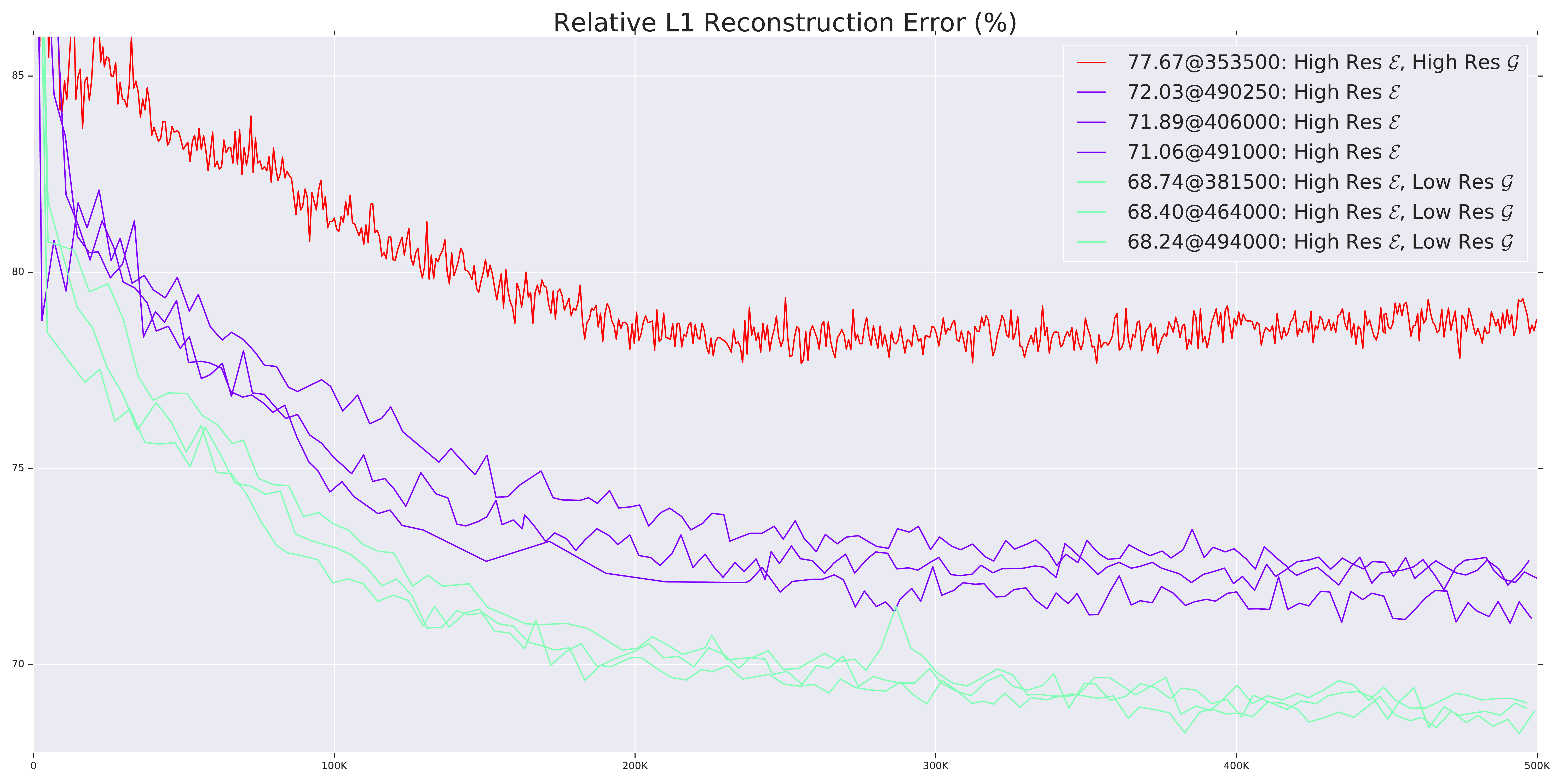}
 \caption{
   Image generation and representation learning curves for high resolution $\enc$ with varying resolution $\gen$ explored in Section~\ref{sec:eval}.
   Legend entries correspond to the following rows in Table~\ref{ablations}:
\textit{High Res $\enc$ (256)},
\textit{Low Res $\gen$ (64)}, and
\textit{High Res $\gen$ (256)}.
}
 \label{fig:curves_highresenc}
\end{figure}

\begin{figure}
\centering
 \includegraphics[width=1.00\textwidth]{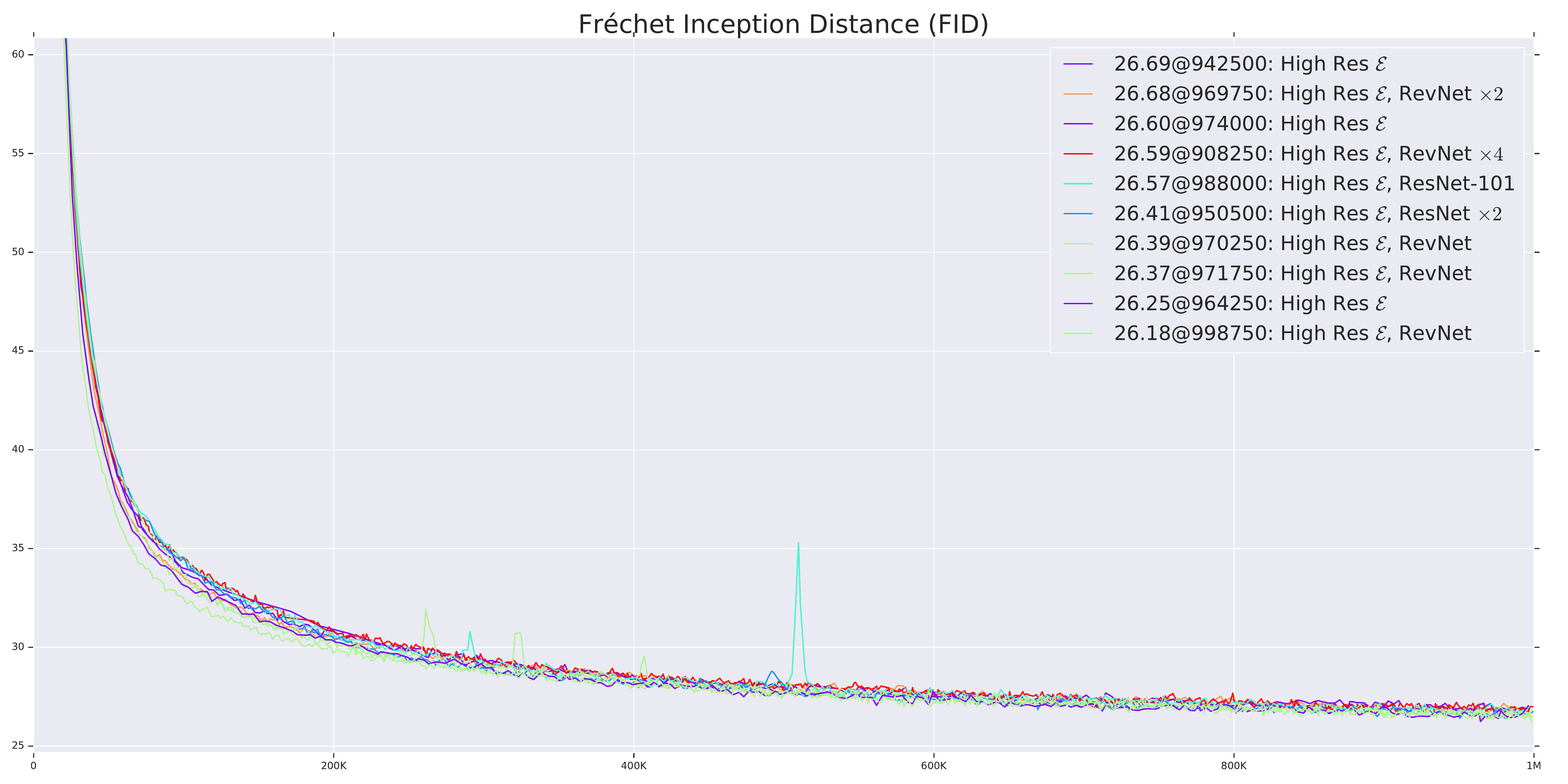}
 \includegraphics[width=1.00\textwidth]{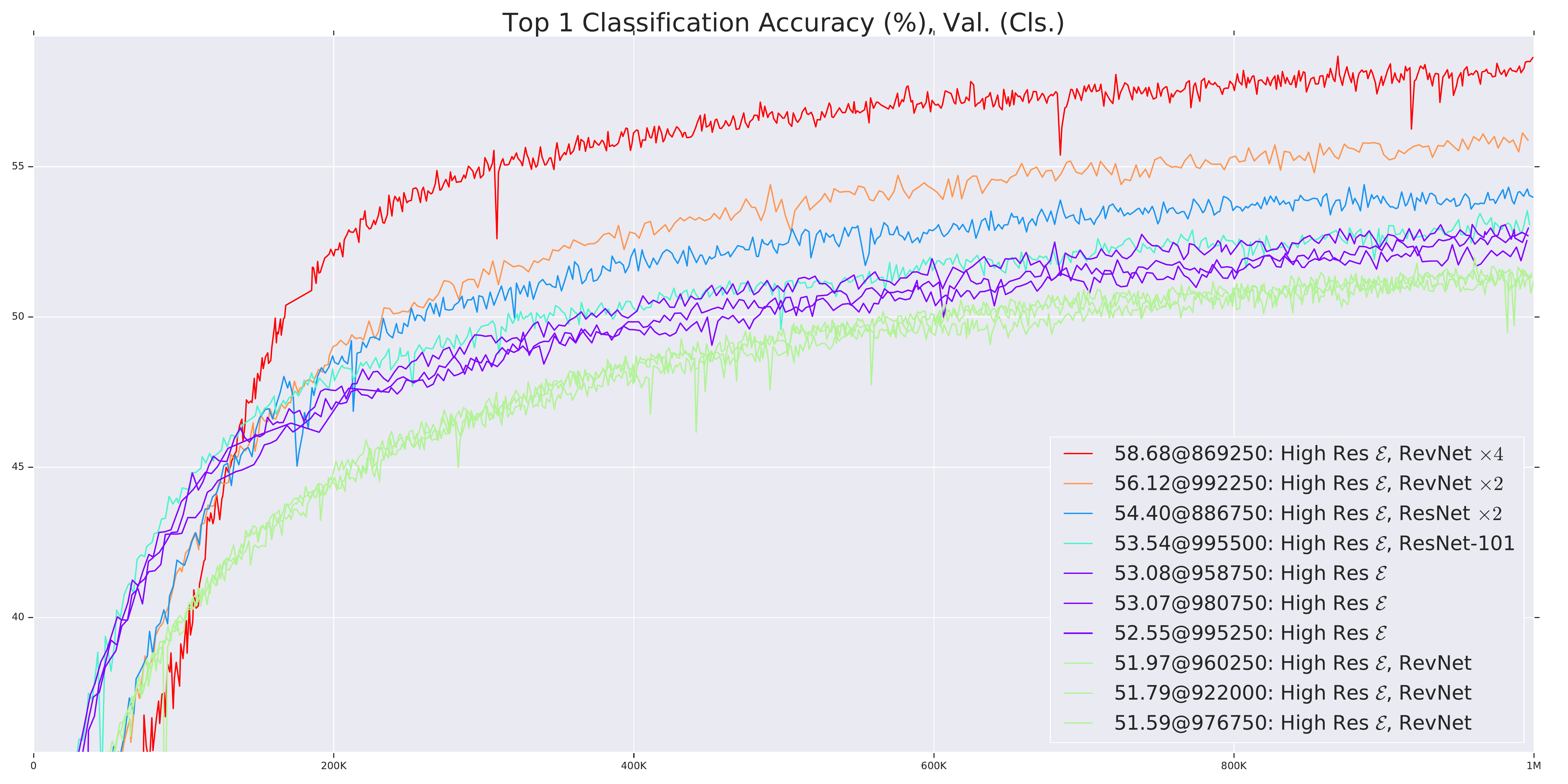}
 \includegraphics[width=1.00\textwidth]{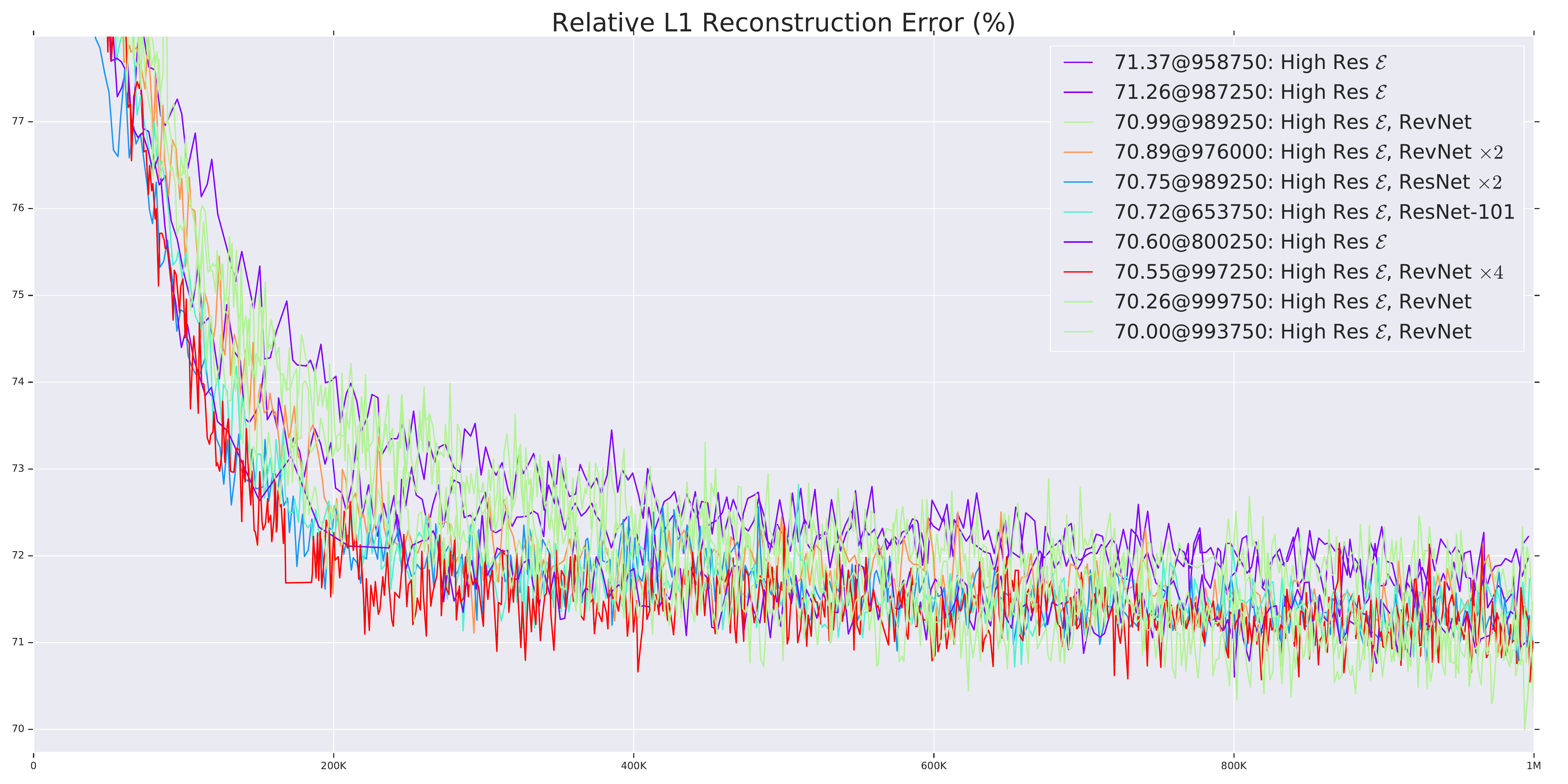}
 \caption{
   Image generation and representation learning curves for the $\enc$ architecture variations explored in Section~\ref{sec:eval}.
   Legend entries correspond to the following rows in Table~\ref{ablations}:
\textit{High Res $\enc$ (256)},
\textit{ResNet-101},
\textit{ResNet $\times 2$},
\textit{RevNet},
\textit{RevNet $\times 2$}, and
\textit{RevNet $\times 4$}.
}
 \label{fig:curves_encarch}
\end{figure}

\begin{figure}
\centering
 \includegraphics[width=1.00\textwidth]{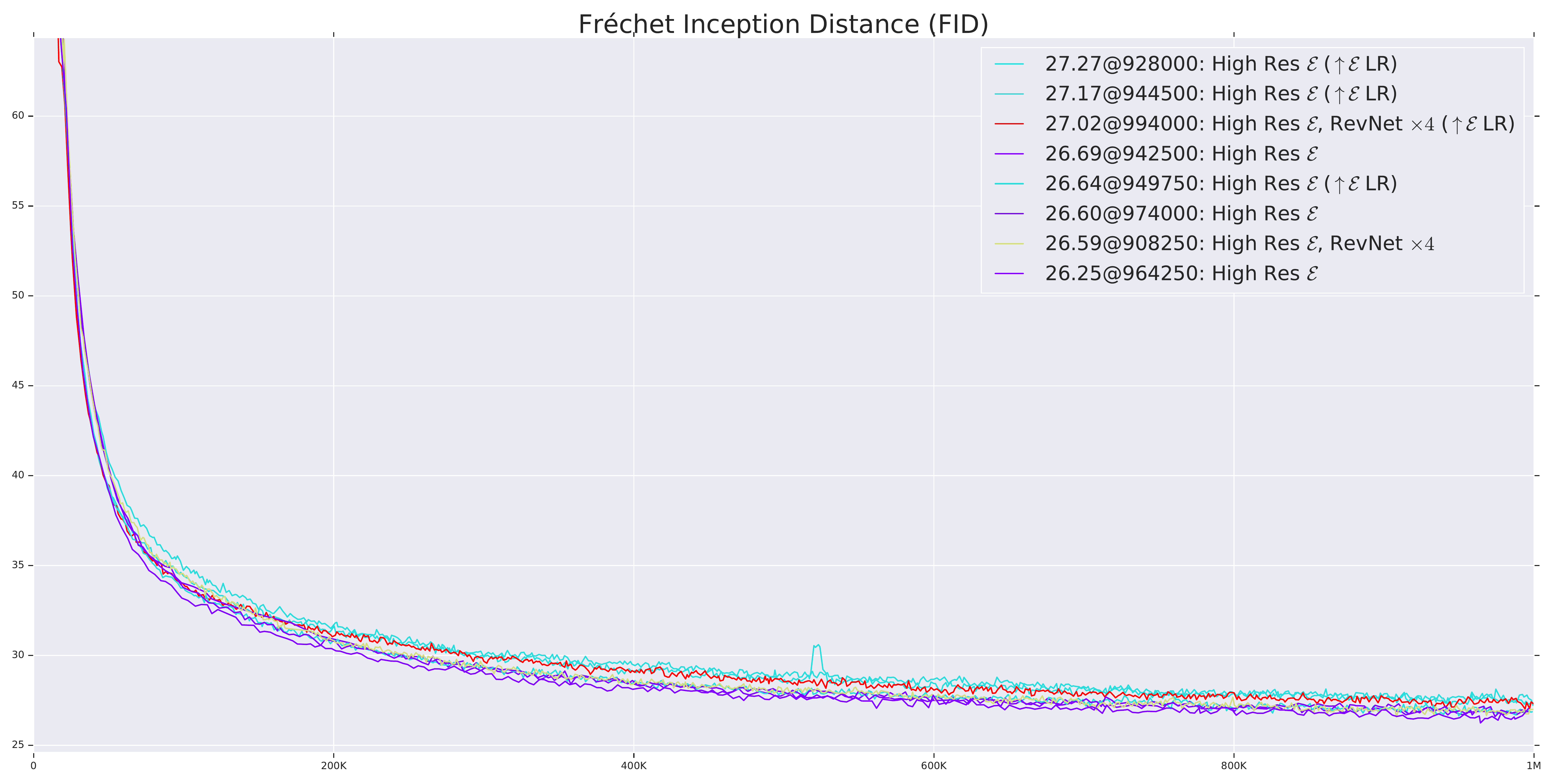}
 \includegraphics[width=1.00\textwidth]{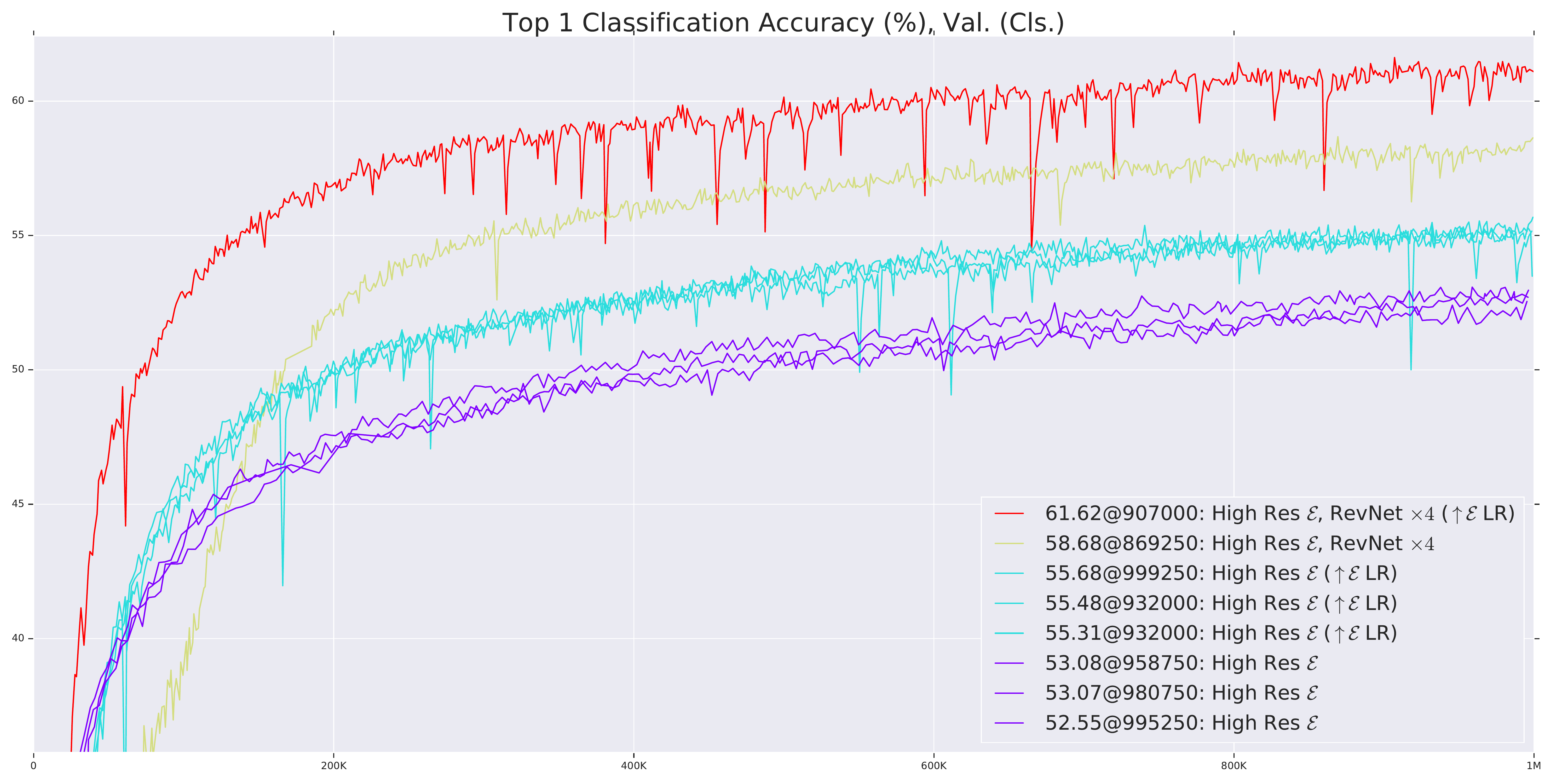}
 \includegraphics[width=1.00\textwidth]{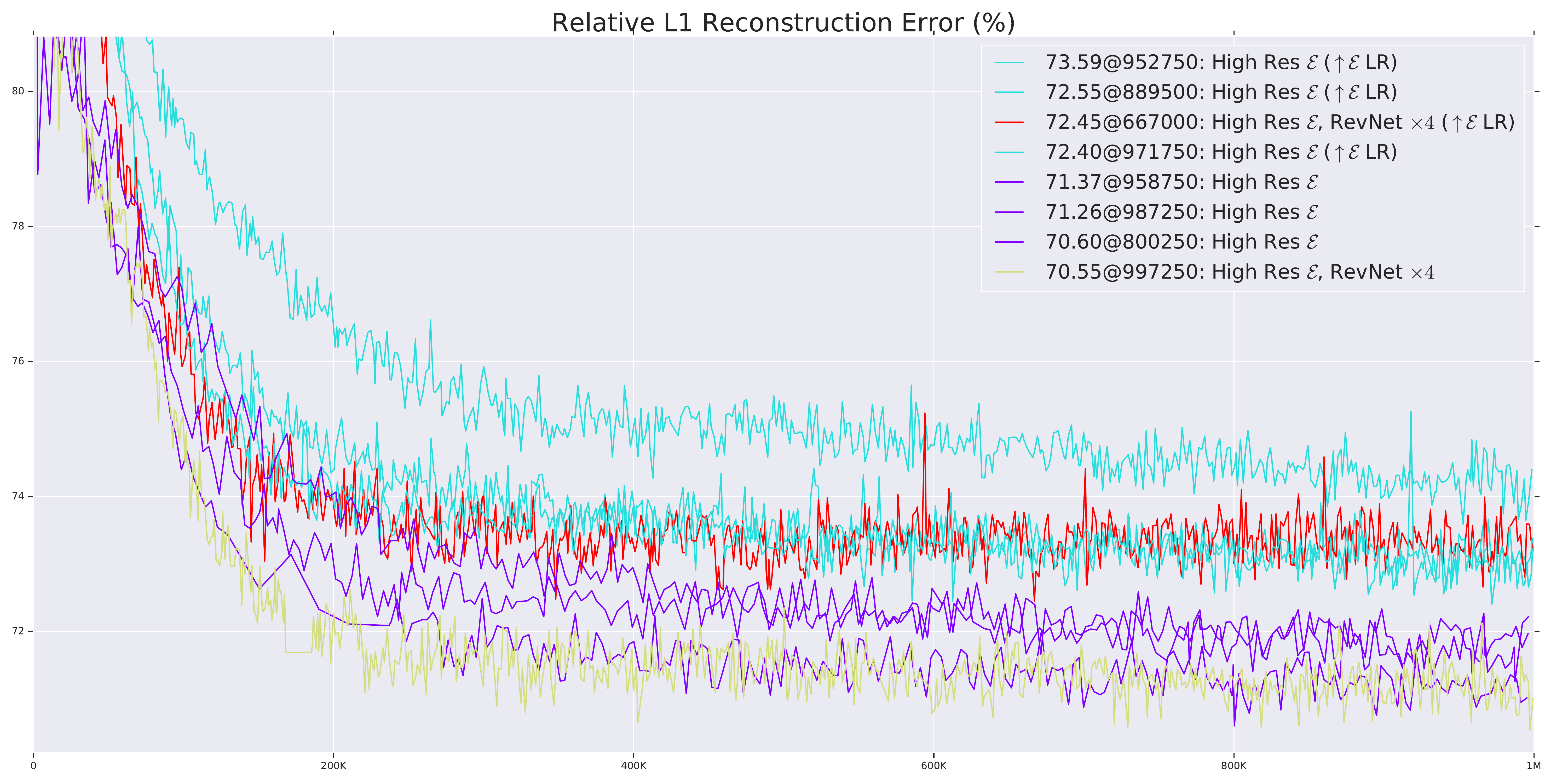}
 \caption{
   Image generation and representation learning curves showing the effect of decoupling the $\enc$ and $\gen$ optimizers to train $\enc$ with $10 \times$ higher learning rate.
   Legend entries correspond to the following rows in Table~\ref{ablations}:
\textit{High Res $\enc$ (256)},
\textit{ResNet ($\uparrow \enc$ LR)},
\textit{RevNet $\times 4$}, and
\textit{RevNet $\times 4$ ($\uparrow \enc$ LR)}.
}
 \label{fig:curves_highlr}
\end{figure}

\begin{samepage}
In this Appendix we present learning curves showing how the image generation and representation learning metrics that we measured evolve throughout training,
as a more detailed view of the results in Section~\ref{sec:eval}, Table~\ref{ablations}.
We include plots for the following results:
\begin{itemize}
\item Image generation (Figure~\ref{fig:curves_gen})
\item Latent distribution $P_\z$ and stochastic $\enc$ (Figure~\ref{fig:curves_latent})
\item Unary loss terms (Figure~\ref{fig:curves_unary})
\item $\gen$ capacity (Figure~\ref{fig:curves_small_g}) \item High resolution $\enc$ with varying resolution $\gen$ (Figure~\ref{fig:curves_highresenc})
\item $\enc$ architecture (Figure~\ref{fig:curves_encarch})
\item Decoupled $\enc$/$\gen$ learning rates (Figure~\ref{fig:curves_highlr})
\end{itemize}
\end{samepage}

\end{appendices}

\end{document}